\documentclass{article}

\usepackage{microtype}
\usepackage{graphicx}
\usepackage{subfigure}
\usepackage{booktabs} %
\usepackage{diagbox}

\usepackage{hyperref}

\usepackage{amsmath,amsfonts,bm}

\def\eqref#1{equation~\ref{#1}}

\def\1{\bm{1}}

\def\rvtheta{{\boldsymbol{\theta}}}

\def\rvm{{\mathbf{m}}}

\def\rvx{{\mathbf{x}}}

\def\rvz{{\mathbf{z}}}

\def\va{{\bm{a}}}
\def\vb{{\bm{b}}}

\def\vs{{\bm{s}}}

\def\vx{{\bm{x}}}
\def\vy{{\bm{y}}}
\def\vz{{\bm{z}}}

\def\mF{{\bm{F}}}
\def\mG{{\bm{G}}}

\def\mW{{\bm{W}}}

\DeclareMathAlphabet{\mathsfit}{\encodingdefault}{\sfdefault}{m}{sl}
\SetMathAlphabet{\mathsfit}{bold}{\encodingdefault}{\sfdefault}{bx}{n}

\newcommand{\E}{\mathbb{E}}

\newcommand{\Cov}{\mathrm{Cov}}

\DeclareMathOperator*{\argmin}{arg\,min}

\usepackage{url}
\usepackage{arydshln}
\usepackage{algorithm}
\usepackage{algorithmic}

\usepackage{booktabs}
\usepackage{multirow}
\usepackage{wrapfig}
\usepackage{adjustbox}

\usepackage[font=small,labelfont=bf,tableposition=top]{caption}
\usepackage{dsfont}

\usepackage{xcolor}
\definecolor{dkgreen}{rgb}{0,0.6,0}
\definecolor{dkred}{rgb}{0.8,0.0,0}
\definecolor{dkblue}{rgb}{0.0,0.0,0.9}
\definecolor{gray}{rgb}{0.5,0.5,0.5}
\definecolor{mauve}{rgb}{0.58,0,0.82}
\definecolor{lightgray}{HTML}{EEEEEE}
\definecolor{dkcyan}{HTML}{008b8b}
\definecolor{earthyellow}{HTML}{F6BE00}
\usepackage{pifont}
\newcommand{\cmark}{\ding{51}}
\newcommand{\xmark}{\ding{55}}

\usepackage{colortbl}

\newcommand{\first}{\cellcolor{blue!40}}
\newcommand{\second}{\cellcolor{blue!20}}
\newcommand{\third}{\cellcolor{blue!10}}
\newcommand{\fourth}{\cellcolor{lightgray}}

\usepackage[accepted]{icml2023}

\usepackage{amsmath}
\usepackage{amssymb}
\usepackage{mathtools}
\usepackage{amsthm}

\usepackage[capitalize,noabbrev]{cleveref}

\theoremstyle{plain}

\theoremstyle{definition}

\theoremstyle{remark}

\usepackage[textsize=tiny]{todonotes}

\icmltitlerunning{Efficient Parametric Approximations of Neural Network FSD}

\begin{document}

\twocolumn[
\icmltitle{Efficient Parametric Approximations of \\ Neural Network Function Space Distance}

\icmlsetsymbol{equal}{*}

\begin{icmlauthorlist}
\icmlauthor{Nikita Dhawan}{uoft,vec}
\icmlauthor{Sicong Huang}{uoft,vec}
\icmlauthor{Juhan Bae}{uoft,vec}
\icmlauthor{Roger Grosse}{uoft,vec,anth}
\end{icmlauthorlist}

\icmlaffiliation{uoft}{University of Toronto}
\icmlaffiliation{vec}{Vector Institute}
\icmlaffiliation{anth}{Anthropic}

\icmlcorrespondingauthor{Nikita Dhawan}{nikita@cs.toronto.edu}

\icmlkeywords{Function Space Distance, Linearization, Continual Learning, Influence Functions}

\vskip 0.3in
]

\printAffiliationsAndNotice{}  %

\begin{abstract}
It is often useful to compactly summarize important properties of model parameters and training data so that they can be used later without storing and/or iterating over the entire dataset. As a specific case, we consider estimating the Function Space Distance (FSD) over a training set, i.e.~the average discrepancy between the outputs of two neural networks. 
We propose a Linearized Activation Function TRick (LAFTR) and derive an efficient approximation to FSD for ReLU neural networks. The key idea is to approximate the architecture as a linear network with stochastic gating. Despite requiring only one parameter per unit of the network, our approach outcompetes other parametric approximations with larger memory requirements. Applied to continual learning, our parametric approximation is competitive with state-of-the-art nonparametric approximations, which require storing many training examples. Furthermore, we show its efficacy in estimating influence functions accurately and detecting mislabeled examples without expensive iterations over the entire dataset.
\end{abstract}

\section{Introduction}
\label{intro}

As machine learning models are trained on increasingly large quantities of data or experience, it can be useful to compactly summarize information contained in a training set. In continual learning, an agent continues interacting with its environment over a long time period --- longer than it is able to store explicitly. 
A natural goal is to avoid overwriting previously learned knowledge as it learns new tasks~\citep{goodfellow2013empirical} while controlling storage costs. 
Even in cases where it is possible to store the entire training set, a compact representation circumvents the need for expensive iterative procedures over the full data. 

We focus on the problem of estimating \emph{Function Space Distance (FSD)} for neural networks: the amount by which the outputs of two networks differ, in expectation over the training distribution. ~\citet{benjamin2018measuring} observed that regularizing FSD over the previous task data is an effective way to prevent catastrophic forgetting. Other tasks such as influence estimation~\citep{bae2022influence}, model editing~\citep{mitchell2021fast}, unlearning~\citep{bourtoule2021machine} and second-order optimization~\citep{amari98,bae2022amortized} have also been formulated in terms of FSD regularization or similar locality constraints.

Methods for summarizing the training data can be categorized as parametric or nonparametric. In the context of preventing catastrophic forgetting, parametric approaches typically store the parameters of a previously trained network, along with additional information about the importance of different directions in parameter space for preserving past knowledge. The canonical example is Elastic Weight Consolidation ~\citep[EWC]{ewc}, which uses a diagonal approximation to the Fisher information matrix. Nonparametric approaches explicitly store, in addition to network parameters, a collection (coreset) of training examples, often optimized directly to be the most important or memorable ones~\citep{rudner2022continual, fromp, frcl}. Currently, the most effective approaches to prevent catastrophic forgetting are nonparametric due to the lack of sufficiently accurate parametric models. However, this advantage comes at the expense of high storage requirements. 

\begin{figure*}[t]
\centering
\begin{minipage}{0.49\textwidth}
        \small
        \includegraphics[width=\linewidth]{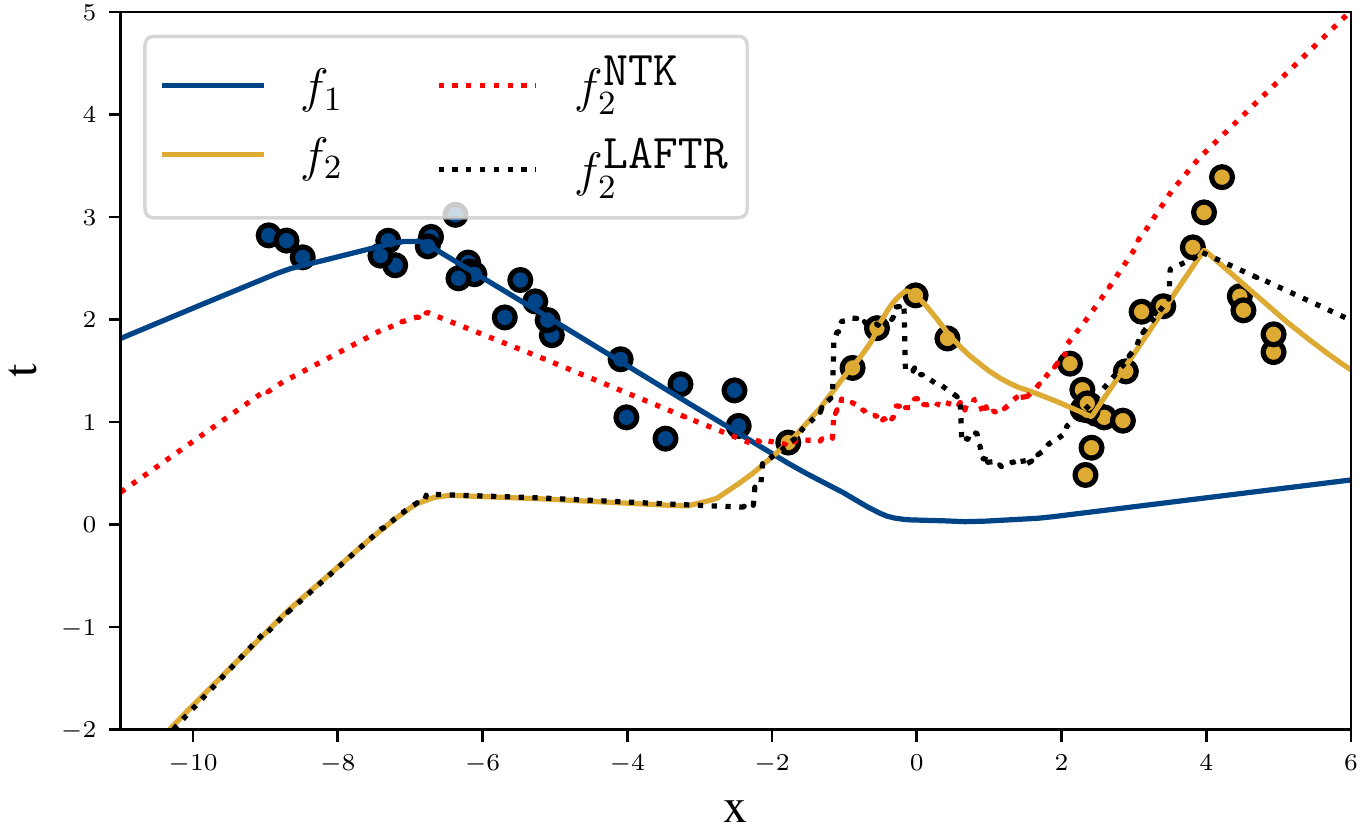}
\end{minipage}
\hfill
\centering
\begin{minipage}{0.49\textwidth}
        \small
        \includegraphics[width=\linewidth]{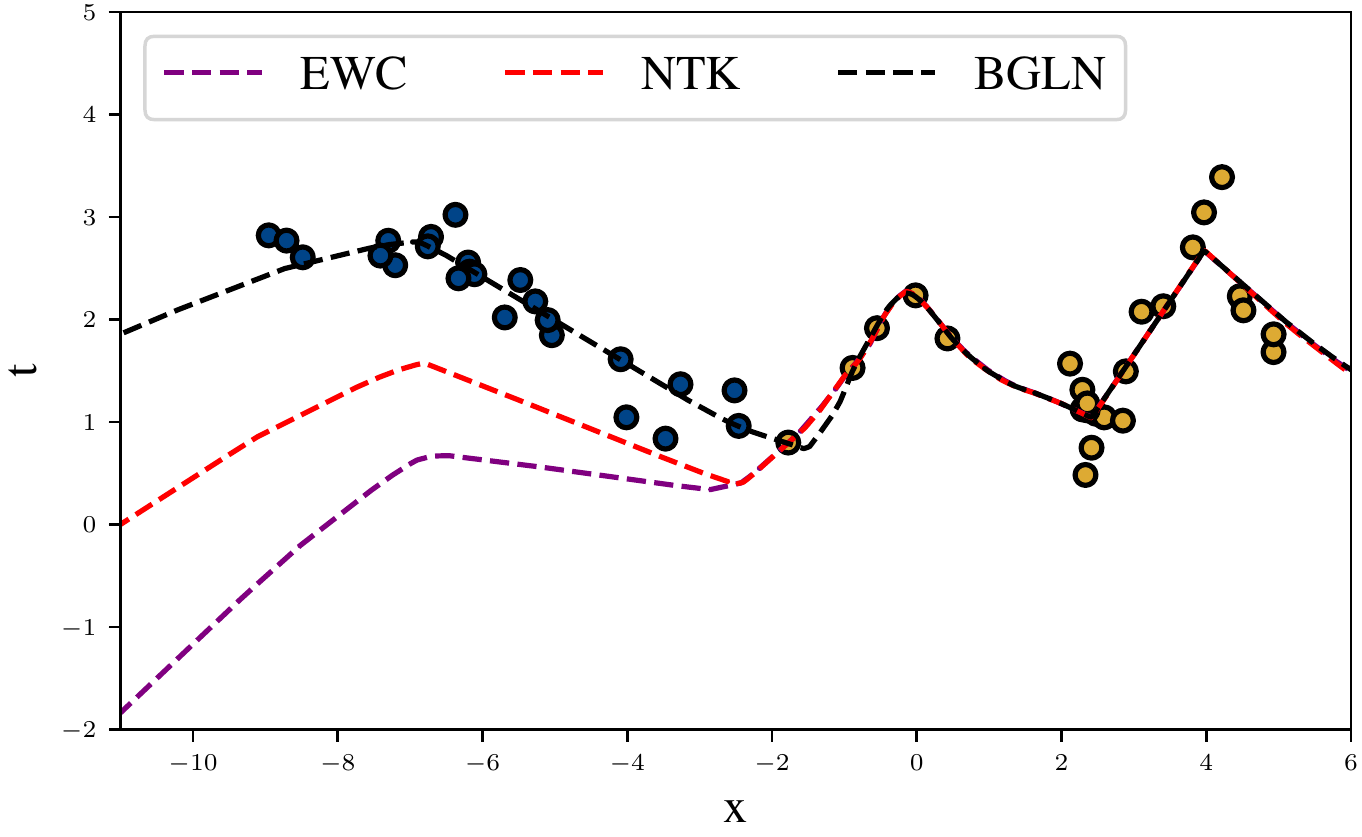}
\end{minipage}
\vspace{-0.2cm}
\caption{Comparison of FSD regularization on a 1-D regression task. \textbf{(Left)} Training sequentially on two tasks ({\color{dkblue}blue} yields $f_1$, then {\color{earthyellow}yellow} yields $f_2$) results in catastrophic forgetting. The LAFTR approximation more closely matches the true function $f_2$ than its NTK approximation does. \textbf{(Right)} BGLN retains performance on task 1 after training on task 2 more accurately than EWC and NTK.}
\label{fig:fsd_1d}
\vspace{-0.4cm}
\end{figure*}

In this paper, we formally formulate neural network FSD estimation and propose novel parametric approximations. 
To motivate our approach, notice that several parametric approximations, like EWC, can be interpreted as a second-order Taylor approximation to the FSD. This leads to a quadratic form involving the Fisher information matrix $\mF_\rvtheta$ or some other metric matrix $\mG_\rvtheta$, where $\rvtheta$ denotes the network parameters. Second-order approximations are practically useful because one can estimate $\mF_\rvtheta$ or $\mG_\theta$ by sampling vectors from a distribution with these matrices as the covariance~\citep{kfac}. Then, tractable probabilistic models can be fit to these samples to approximate the corresponding distribution. Unfortunately, these tend to be inaccurate for continual learning compared to nonparametric approaches. We believe the culprit is the second-order Taylor approximation: we show in several examples that even the exact second-order Taylor approximation can perform poorly in terms of average classification accuracy and backward transfer in sequentially learned tasks. Since such an approximation can be interpreted as network linearization~\citep{GrosseNNTDChapter4}, this finding is consistent with a recent line of results that find linearized approximations of neural networks to be an inaccurate model of their behavior~\citep{seleznova2022analyzing, seleznova2022neural, hanin2019finite, bai2020taylorized, huang2020dynamics}. In Section~\ref{background}, we present this network linearization perspective of some existing approaches for regularization in function space.

Our method, based on a \emph{Linearized Activation Function TRick (LAFTR)}, does \emph{not} make a second-order Taylor approximation in the parameter space, and hence is able to capture nonlinear interactions between parameters of the network. Specifically, it linearizes each step of the network's forward pass with respect to its inputs. In the case of ReLU networks, our approximation yields a linear network with stochastic gating, which we refer to as the \emph{Bernoulli Gated Linear Network (BGLN)}. 
We derive a stochastic and a deterministic estimate of FSD, both of which rely only on the first two moments of the data. This allows the application of our methods in different scenarios where stochasticity is or isn't desirable. 

We evaluate our BGLN approximation in the contexts of continual learning and influence function estimation. Our method significantly outperforms previous parametric approximations despite being much more memory-efficient. For continual learning tasks, our method is competitive with nonparametric approaches. For influence function estimation tasks, it closely matches an oracle estimate of a network's loss after a data point is removed, but without having to iterate over the whole dataset.

The key contributions and findings of this work are:
\begin{itemize}
\vspace{-0.2cm}
\setlength\itemsep{-0.1em}
    \item We introduce LAFTR, an idealized FSD approximation, which improves over parameter space linearization by capturing nonlinear interactions between weights in different layers. 
    \item We propose the Bernoulli Gated Linear Network (BGLN), an efficient parametric FSD approximation for ReLU networks based on LAFTR which stores only aggregate statistics of the data and the activations.
    \item In continual learning, BGLN outcompetes state-of-the-art methods on sequential MNIST and CIFAR100 tasks, with significantly lower memory requirements than nonparametric methods. 
    \item For influence function estimation, BGLN efficiently approximates the effect of removing a single data point without iterating over or storing the full dataset. 

\end{itemize}

\section{Background}
\label{background}

Let $\vz = f(\rvx, \rvtheta)$ denote the function computed by a neural network, which takes in inputs $\rvx$ and parameters $\rvtheta$.
Consistent with prior works, we use FSD to refer to the expected output space distance\footnote{Note that we use the term \emph{distance} throughout since we focus on Euclidean distance in our derivation. However, other metrics like KL divergence can also be used, as shown in Section~\ref{expts}.} $\rho$ between the outputs of two neural networks~\citep{benjamin2018measuring,GrosseNNTDChapter4,bae2022amortized} with respect to the training distribution, as defined in equation~\ref{fsd_def}. When the population distribution is inaccessible, the empirical distribution is often used as a proxy:
\begin{align}
    \label{fsd_def}
    D(\rvtheta_0, \rvtheta_1, p_{{\text{data}}}) &= \mathbb{E}_{\rvx \sim p_{{\text{data}}}} [\rho(f(\rvx, \rvtheta_0), f(\rvx, \rvtheta_1))] \\
    \label{emp_fsd}
    &\approx \frac{1}{N} \sum_{i=1}^N \rho(f(\rvx^{(i)}, \rvtheta_0), f(\rvx^{(i)}, \rvtheta_1)),
\end{align} 
where $p_{\text{data}}$ is the data-generating distribution.
Constraining the FSD term has been successful in preventing catastrophic forgetting~\citep{benjamin2018measuring}, computing influence functions~\citep{bae2022influence}, training teacher-student models~\citep{hinton2015distilling}, and fine-tuning pre-trained networks~\citep{jiang2019smart,mitchell2021fast}. Natural choices for $\rho$ are Euclidean distance for networks trained using mean-squared error (e.g. regression) and KL divergence for those trained with cross-entropy loss (e.g. classification). 

Consider the continual learning setting as a motivating example. Common benchmarks~\citep{normandin2021sequoia} involve sequentially learning tasks $t \in \{1, \dots, T\}$, using loss function $\mathcal{L}$ and a penalty on the FSD between the parameters $\rvtheta$, and the parameters $\{\rvtheta_i\}$ fit to previous tasks. The penalty is computed over the previously seen data distribution $p_i$ and then scaled by a hyperparameter $\lambda_{\text{FSD}}$:
\vspace{-0.2cm}
\begin{equation}
\label{cl_eq}
    \rvtheta_t = \arg \min_{\rvtheta} \mathcal{L}(\rvtheta) + \lambda_{\text{FSD}} \sum_{i=1}^{t-1} D(\rvtheta, \rvtheta_i, p_i).
\end{equation}
Continuing with the notation in equation~\ref{emp_fsd}, one way to regularize the FSD is to store the training set and explicitly evaluate the network outputs using both $\rvtheta_0$ and $\rvtheta_1$ (perhaps on random mini-batches). However, this has the drawbacks of having to store and access the entire training set throughout training (precisely the thing continual learning research tries to avoid) and necessarily estimating FSD stochastically. Instead, we would like to compactly summarize information about the training set or distribution.

Many (but not all) practical FSD approximations are based on a second-order Taylor approximation: 
\begin{equation}
\label{taylor_approx}
\begin{aligned}
    D(\rvtheta_0, \rvtheta_1, p_{{\text{data}}}) 
    \approx  \frac{1}{2} (\rvtheta_1 - \rvtheta_0)^\top \mG_{\rvtheta} (\rvtheta_1 - \rvtheta_0),
\end{aligned}
\end{equation}
where $\mG_{\rvtheta} = \nabla_{\rvtheta}^2 D(\rvtheta_0, \rvtheta, p_{{\text{data}}})$ is the corresponding Hessian. 
In the case where the network outputs parametrize a probability distribution and $\rho$ corresponds to KL divergence, $\mG_{\rvtheta}$ reduces to the more familiar Fisher information matrix $F_{\rvtheta} = \mathbb{E}_{\rvx \sim p_{{\text{data}}}, \mathbf{y} \sim P_{\mathbf{y}|\mathbf{x}}(\rvtheta)}[\nabla_{\rvtheta} \log p(\mathbf{y}|\rvtheta,\rvx) \nabla_{\rvtheta} \log p(\mathbf{y}|\rvtheta,\rvx)^\top]$, where $P_{\mathbf{y}|\mathbf{x}}(\rvtheta)$ represents the model's predictive distribution over $\mathbf{y}$. 
It is possible to sample random vectors in parameter space whose covariance is $\mG_{\rvtheta}$~\citep{martens2012estimating,grosse2016kronecker,GrosseNNTDChapter4} and some parametric FSD approximations work by fitting simple statistical models to the resulting distribution. For instance, assuming all coordinates are independent gives a diagonal approximation~\citep{ewc}, and more fine-grained independence assumptions between network layers yield a Kronecker-factored approximation~\citep{kfac, kfac_cl}. In practice, instead of sampling vectors whose covariance is $\mG_{\rvtheta}$, many works use the empirical gradients during training, whose covariance is the empirical Fisher matrix. We caution the reader that the empirical Fisher matrix is less well motivated theoretically and can result in different behavior~\citep{kunstner2019limitations}.

\section{A Parametric Estimate with LAFTR}
\label{method}

We introduce and apply \textbf{LAFTR} (Linearized Activation Function TRick) to linear ReLU networks and propose \textbf{BGLN} (Bernoulli Gated Linear Network) which approximates a given model architecture as a linear network with stochastic gating.
While it is applicable to different architectures, we first explicitly derive our approximation for multilayer perceptrons (MLPs) with $L$ fully-connected layers and ReLU activation function $\phi$. We also discuss its generalization to convolutional networks and empirically evaluate the same in Section~\ref{expts}. 

For MLPs with inputs $\rvx$ drawn from $p_{\text{data}}$, layer $l$ weights and biases $(\mW^{(l)}, \vb^{(l)})$, and outputs $\vz$, the computation of preactivations and activations at each layer is recursively defined as follows:
\begin{equation}
\label{model_def}
    \vs^{(l)} = \mW^{(l)} \va^{(l-1)} + \vb^{(l)}, ~ \va^{(l)} = \phi(\vs^{(l)}) 
\end{equation} 
with $\va^{(0)} = \vx$, and $\vs^{(L)} = \vz$. We denote $\vz_0$ and $\vz_1$ to be samples of the output distribution obtained with parameters $\rvtheta_0$ and $\rvtheta_1$, respectively. 

\subsection{Linearized Activation Function TRick}

Given parameters $\rvtheta_0$ and $\rvtheta_1$
of two networks, we linearize \emph{each step of the forward pass} around its value under $\rvtheta_0$. For an MLP that alternates between linear layers and non-linear activation functions, the linear transformations are unmodified while the activation functions are replaced with a first-order Taylor approximation around their inputs. Hence, the network's computation becomes linear in $\rvx$ (but, importantly, remains nonlinear in $\rvtheta$). Let $(\mW_i^{(l)}, \vb_i^{(l)})$ denote the weights and biases of layer $l$ in network $i$.
\begin{align}
    \label{s0}
    \vs_0^{(l)} &= \mW_0^{(l)} \va^{(l-1)}_0 + \vb_0^{(l)} \\
    \label{a0}
    \va_0^{(l)} &= \phi(\vs_0^{(l)}) \\
    \label{s1}
    \vs_1^{(l)} &= \mW_1^{(l)} \va_1^{(l-1)} + \vb_1^{(l)} \\
    \label{a1}
    \va_1^{(l)} &= \phi(\vs_0^{(l)}) +  \phi'(\vs_0^{(l)}) \odot (\vs_1^{(l)} - \vs_0^{(l)}),
\end{align}
where $\phi'$ is the derivative of the activation function.

We define some additional notation for differences between preactivations and activations. For $\Delta \vs^{(l)} = \vs_1^{(l)} - \vs_0^{(l)}$, $\Delta \va^{(l)} = \va_1^{(l)} - \va_0^{(l)}$, $\Delta \mW^{(l)} = \mW_1^{(l)} - \mW_0^{(l)}$, and $\Delta \vb^{(l)} = \vb_1^{(l)} - \vb_0^{(l)}$, we have the following formulae:
\begin{align}
    \label{delta_s}
    \Delta \vs^{(l)} &= \Delta \mW^{(l)} \va_0^{(l-1)} + \mW_1^{(l)} \Delta \va^{(l-1)} + \Delta \vb^{(l)}  \\
    \label{delta_a}
    \Delta \va^{(l)} &= \phi'(\vs_0^{(l)}) \odot \Delta \vs^{(l)}.
\end{align}
Here, base cases are $\va_0^{(0)} = \rvx$ and $\Delta \va^{(0)} = 0$.
Written this way, the $\Delta$ terms at each step of computation rely linearly on corresponding terms from the immediately preceding step, making LAFTR conducive to implementation via propagation of the base cases through the network. 
Observe that with $\mW_0$ held fixed, the model parametrized using $\mW_1$ is a linear network, i.e.~the network's exact outputs are a linear function of its inputs. 

There are two significant differences between LAFTR and parameter space linearization (referred to as NTK~\citep{ntk} henceforth), which confer an advantage to the former. First, our linearization is with respect to inputs instead of parameters~\citep{lee2019wide}, hence capturing nonlinear interactions between the parameters in different layers. Second, the only computations that introduce linearization errors into our approximation are those involving activation functions; hence, our method is exact for linear networks, whereas NTK is only approximate. We note that linear networks are commonly used to model nonlinear training dynamics of neural networks~\citep{saxe2013exact}. Hence, we regard our weaker form of linearity as a significant advantage over Taylor approximations in parameter space. We note that LAFTR can be extended directly to other types of linear layer, such as convolution layers.

In the following two sections, we use the intuition above to motivate two probabilistic approximations which enable a memory-efficient implementation of this algorithm. One is to approximate preactivation signs and the other is for the input data distribution.

\subsection{Bernoulli Gating}
\label{bernoulli_mle}
In the specific case of ReLU networks, observe that LAFTR depends on the training data only through the signs of preactivations, which we denote as a mask $\rvm = \mathds{1}\{\vs>0\}$, since $\phi(\vs) = \rvm \odot \vs$ and $\phi'(\vs) \odot \Delta \vs = \rvm \odot \Delta \vs$. Here, $\phi'$ is the derivative of the ReLU function, given by $\phi'(\vs) = \mathds{1}\{\vs>0\}$.
We approximate $\rvm$ as a vector of independent Bernoulli random variables and fit its mean vector $\boldsymbol{\mu}$ using maximum likelihood estimation (i.e., computing the fraction of times the unit is activated). We can accordingly rewrite equations \ref{a0} and \ref{delta_a} as $\va_0^{(l)} = \rvm^{(l)} \odot \vs_0^{(l)}$ and $\Delta \va^{(l)} = \rvm^{(l)} \odot \Delta \vs^{(l)}$, respectively, where $\odot$ denotes element-wise multiplication and $\rvm^{(l)} \sim Ber(\boldsymbol{\mu}^{(l)})$. For efficiency, we compute the activation statistics over the last epoch of training. %
We call this approximation the Bernoulli Gated Linear Network (BGLN). 

Note that this gating technique is not specific only to MLPs. It can be implemented in ReLU convolutional networks by replacing activations with Bernoulli random variables.

\subsection{Propagating Moments of the Activations}
\label{gauss_inputs}
A key insight enabling efficient computation is that when the output distance $\rho$ is chosen to be (squared) Euclidean distance, the FSD depends only on the first and second moments of the output difference $\Delta \vz := \vz_1 - \vz_0$: 
\begin{equation}
\label{out_moments}
    \E \left[\tfrac{1}{2}||\Delta \vz||^2 \right] 
   = \tfrac{1}{2} ||\E [\Delta \rvz]||^2 + \tfrac{1}{2} \textit{tr }( \Cov (\Delta \rvz)).
\end{equation}
We compute these terms recursively by propagating the first two moments of $\va_0^{(l)}$ and $\Delta \va^{(l)}$ through the network. Using equations \ref{delta_s} and \ref{delta_a}, we obtain the following equations for the first moments (see Appendix~\ref{app:bgln-d} for analogous equations for second moments):
\begin{align*}
    &\E [\vs_0^{(l)}] = \mW_0^{(l)} \E [\va^{(l-1)}_0] + \vb_0^{(l)} \\
    &\E [\va_0^{(l)}] = \boldsymbol{\mu}^{(l)} \odot \E [\vs_0^{(l)}] \\
    &\E [\Delta \vs^{(l)}] = \Delta \mW^{(l)} \E [\va_0^{(l-1)}] + \mW_1^{(l)} \E [\Delta \va^{(l-1)}] + \Delta \vb^{(l)}  \\
    &\E [\Delta \va^{(l)}] = \boldsymbol{\mu}^{(l)} \odot \E [\Delta \vs^{(l)}]
\end{align*}
Hence, LAFTR-based FSD approximation depends only on the first two moments of the data. This view leads to two insights: (1) we can store the first two moments of the data instead of a coreset and compute a deterministic FSD estimate and (2) we can approximate the data distribution as a multivariate Gaussian parameterized by its moments and compute an unbiased stochastic estimate of the LAFTR-based FSD by sampling from it.

\subsection{BGLN-D and BGLN-S}
\label{bgln}
The most straightforward way to use the BGLN approximation is to draw Monte Carlo samples of the random variables. When only the first and second moments matter, we are free to assume Gaussianity of the inputs. We denote this method BGLN-S (S for ``stochastic''). This is sufficient in situations where FSD is used as a regularization term in stochastic gradient-based optimization (as we do in our continual learning experiments). In other situations, it is advantageous to have a deterministic computation; for instance, optimization with nonlinear conjugate gradient requires a deterministic objective. In the case of Euclidean distance as the output space metric, we can exactly compute the BGLN approximation by propagating the first and second moments of all random variables through the forward pass, as described in Section \ref{gauss_inputs}. We call this deterministic estimator BGLN-D. \textbf{BGLN-S} is outlined in Algorithm \ref{stoch_alg} and \textbf{BGLN-D} in Algorithm \ref{determ_alg}. We describe the analogous BGLN-S computations for convolutional networks in Appendix \ref{app:conv}.

We present the above algorithms such that they depend on storing the mean and covariance of the data. Note that storing moments of the data requires less memory than storing sufficient subsets of the data itself in several practical settings demonstrated in Table \ref{memory_gain}. We can further reduce the memory requirement by approximating the covariance matrix as a diagonal matrix, i.e., using only the variance of each dimension of the inputs. This is equivalent in cost to storing two data points per task. We empirically investigate the effect of this approximation on continual learning benchmarks. Furthermore, capturing the expected FSD over the training distribution in a single term, via a single forward pass, is far less computationally expensive than iterative alternatives.

\subsection{Class-conditional Estimates}
\label{class-cond}

In the classification setting, it is also possible to a extend BGLN to a more fine-grained, class-conditional approximation. In particular, we can fit a mixture model for our probabilistic approximations, with one component per class. In this case, each class has its own associated input moments and Bernoulli mean parameters. The lower memory cost of our method allows for this when number of classes is not too large. We refer to theis variant as BGLN-CW. As expected, it boosts performance in our continual learning experiments at the expense of a slightly higher memory requirement, as shown in Tables~\ref{table:avg_mnist},~\ref{table:bwt_mnist} and~\ref{table:kl_cifar}.

\begin{algorithm}[tb]
    \caption{A stochastic version of BGLN (BGLN-S)}
    \label{stoch_alg}
    \begin{algorithmic}[1]
        \REQUIRE{$\mathbb{E}[\rvx], \mathrm{Cov}(\rvx), \{\mW, \vb\}_1^{L}, \{\boldsymbol{\mu}\}_1^{L-1}$}
        \STATE {$\mathbb{E} [\va_0], \mathbb{E} [\va_1] \gets \mathbb{E}[\rvx]$} 
        \STATE {$\mathrm{Cov} (\va_0), \mathrm{Cov} (\va_1) \gets \mathrm{Cov}(\rvx)$} 
        \STATE {$\va_0, \va_1 \sim \mathcal{N}(\mathbb{E}[\va_0], \mathrm{Cov}(\va_0))$} \COMMENT{sample inputs}
        \STATE {$\Delta \va \gets 0$}
        \FOR{$l \gets 1$ to $L-1$}  
            \STATE {$\rvm \sim Ber(\boldsymbol{\mu^{(l)}})$} \COMMENT{sample Bernoullis}
            \STATE {$\vs_0 \gets \mW_0^{(l)} \va_0 + \vb_0^{(l)}$} 
            \STATE {$\Delta \vs \gets \Delta \mW^{(l)} \va_0 + \mW_1^{(l)}  \Delta \va + \Delta \vb^{(l)}$}
            \STATE {$\va_0 \gets \rvm \odot \vs_0$}  \COMMENT{stochastic gating}
            \STATE {$\Delta \va \gets \rvm \odot \Delta \vs$}  
            \STATE {$\vs_1 \gets \mW_1^{(l)} \va_1 + \vb_1$} 
            \STATE {$\va_1 \gets \va_0 + \Delta \va$} \COMMENT{linearized activations}
        \ENDFOR
        \STATE {$\Delta \vz \gets \Delta \mW^{(L)} \va_0 + \mW_1^{(L)}  \Delta \va + \Delta \vb^{(L)}$} 
        \STATE \textbf{return } {$\frac{1}{2}||\Delta \vz||^2$}
    \end{algorithmic}
\end{algorithm}
\section{Related Works}
\label{rel_works}

Several works~\citep{benjamin2018measuring, fromage, bae2022amortized} have highlighted the importance of measuring meaningful distances between neural networks.~\citet{benjamin2018measuring} contrast training dynamics in parameter space and function space and observe that function space distances are often more useful than, and not always correlated with, parameter space distances.~\citet{bae2022amortized} propose an Amortized Proximal Optimization (APO) scheme that regularizes an FSD estimate to the previous iterate for second-order optimization. Natural gradient descent~\citep{amari95, amari98} can also be interpreted as a steepest descent method, using a second-order Taylor approximation to the FSD~\citep{Pascanu2014RevisitingNG}.

\citet{cl_review, de2021continual, ramasesh2020anatomy, normandin2021sequoia} have reviewed and surveyed the challenge of catastrophic forgetting in continual learning, along with benchmarks and metrics to evaluate different methods. Parametric methods focus on different approximations to the weight space metric matrix, like diagonal~\citep[EWC]{ewc} or Kronecker-factored~\citep[OSLA]{kfac_cl}. As described in Section~\ref{background}, we interpret these as second-order Taylor approximations to the FSD with further structured approximations to the Hessian. Several methods are motivated as approximations to a posterior Gaussian distribution in a Bayesian setting~\citep{ebrahimi2019uncertainty}, for instance through a variational lower bound~\citep{vcl} or via Gaussian process inducing points~\citep{var_gp}. Non-parametric methods~\citep{var_gp, frcl, fromp, rudner2021continual, rudner2022continual, kirichenko2021task} usually employ some form of experience replay of stored or optimized data points. Some of these methods~\citep{fromp} can also be related to the Neural Tangent Kernel~\citep[NTK]{ntk}, or in other words, network linearization.~\citet{doan2021theoretical} directly study forgetting in continual learning in the infinite width NTK regime.~\citet{mirzadeh2022wide} further study the impact of network widths on forgetting.

In this paper, we also examine influence functions~\citep{cook1979influential, hampel} which is another application that involves the FSD between networks. Influence functions are a classical robust statistics technique that has since been used in machine learning~\citep{koh2017understanding}.~\citet{bae2022influence} formally study influence functions in neural networks and show that they approximate an objective called the proximal Bregman response function (PBRF). This approximation depends on a FSD term that is typically computed by iterating through the full training dataset. 

\section{Experiments}
\label{expts}

We empirically assess the effectiveness of LAFTR (our idealized method) and the BGLN (our practical algorithm) in approximating FSD as well as their usefulness for downstream tasks: continual learning and influence function estimation.
The experiments investigate the following questions: 
\begin{itemize}
\vspace{-0.3cm}
\setlength\itemsep{-0.2em}
    \item Can LAFTR outperform NTK in approximating FSD?
    \item Does LAFTR improve performance and memory cost on continual learning benchmarks relative to existing methods?
    \item How do choice of output space metric $\rho$, or the use of the Gaussian input and the Bernoulli activation approximations, impact empirical performance?
    \item Can the BGLN perform competitively with iteration-based influence function estimators without requiring iteration over the dataset?
\end{itemize}

\begin{table}[t]
\centering
\scriptsize
\begin{center}
\resizebox{\columnwidth}{!}{%
\begin{tabular}{lcc}
\toprule
\multicolumn{1}{l}{\bf Method}  &\multicolumn{1}{c}{\bf Split MNIST} &\multicolumn{1}{c}{\bf Permuted MNIST}\\ %
\midrule
\underline{Nonparametric} \\
\textbf{VCL (coreset)} & \third $98.40$ & \third $95.50$\\ %
\textbf{VAR-GP (coreset)} & $90.57\pm1.06$ & \first $\mathbf{97.20\pm0.08}$\\ %
\textbf{FROMP (coreset)} & \second $99.00\pm0.04$ & \fourth $94.90\pm0.04$\\ %
\textbf{S-FSVI (coreset)} & \second $99.54\pm0.04$ & \third $95.76\pm0.02$\\ %
\textbf{NTK (coreset)} & \second $99.50\pm0.09$ & \second $96.46\pm0.11$\\ %
\textbf{BGLN-S (coreset)} & \second $99.50\pm0.03$ & \second $96.36\pm0.13$\\ %
\midrule
\underline{Parametric} \\
\textbf{EWC} & $63.10$ & $84.00$\\ %
\textbf{OSLA} & $80.56$ & \third $95.73$\\ %
\textbf{VCL} & \fourth $97.00$ & $87.50\pm0.61$\\ %
\textbf{BGLN-D} & \first $99.72\pm0.03$ & \second $96.03\pm0.20$\\ %
    \textbf{BGLN-D-CW} & \first $\mathbf{99.78\pm0.02}$ & \first $96.85\pm0.02$\\ %
\textbf{BGLN-S} & \first $99.64\pm0.04$ & \second $96.36\pm0.12$\\ %
\textbf{BGLN-S-CW} & \first $99.77\pm0.05$ & \first $96.99\pm0.07$\\ %
\textbf{BGLN-D-Var} & \first $99.64\pm0.04$ & \third $94.98\pm0.18$ \\ %
\textbf{BGLN-S-Var} & \second $99.50\pm0.03$ & \second $96.36\pm0.13$ \\ %
\hline
\end{tabular}%
}
\vspace{-0.5cm}
\end{center}
\caption{Average accuracies of nonparametric and parametric approaches on Split and Permuted MNIST datasets.
}
\label{table:avg_mnist}
\vspace{-0.3cm}
\end{table}

\begin{table}[t]
\centering
\scriptsize
\begin{center}
\resizebox{\columnwidth}{!}{
\begin{tabular}{lcc}
\toprule
\multicolumn{1}{l}{\bf Method}  &\multicolumn{1}{c}{\bf Split MNIST} &\multicolumn{1}{c}{\bf Permuted MNIST}\\ %
\midrule
\textbf{FROMP} & \fourth $-0.50\pm0.20$ & \third $-1.00\pm0.10$\\ %
\textbf{S-FSVI} & \third $-0.21\pm0.06$ & \second $-0.65\pm0.21$\\ %
\midrule
\textbf{BGLN-S} & \first $\mathbf{-0.04\pm0.03}$ & \first $-0.41\pm0.08$\\ %
\textbf{BGLN-D} & \first $-0.09\pm0.04$ & \second $-0.56\pm0.04$\\ %
\textbf{BGLN-S-CW} & \second $-0.18\pm0.06$ & \first $\mathbf{-0.37\pm0.14}$\\ %
\textbf{BGLN-D-CW} & \first $-0.07\pm0.07$ & \fourth $-1.17\pm0.07$\\ %
\hline
\end{tabular}
}
\vspace{-0.5cm}
\end{center}
\caption{Backward transfer on Split and Permuted MNIST. Higher is better.}
\vspace{-0.3cm}
\label{table:bwt_mnist}
\end{table}

\subsection{Comparing FSD Estimators}
To conduct further empirical analysis of our methods' estimation and minimization of the true (empirical) FSD, we use tasks and models from standard continual learning settings which are prone to forgetting. These include Split MNIST, Permuted MNIST and Split CIFAR100~\citep{fromp, rudner2022continual}.
 In addition to directly evaluating the continual learning performance, we also use a collection of networks trained in the course of this experiment (with varying hyperparameter settings) to directly evaluate the accuracy of the FSD estimates. Specifically, we vary the learning rate and the number of training iterations, and consider the set of trained networks that result; on each pair of these networks, we compare the FSD estimates against the true empirical FSD computed using the full training set.

\begin{figure*}[t]
\begin{minipage}{0.32\textwidth}
    \small
    \centering
    \includegraphics[width=\linewidth]{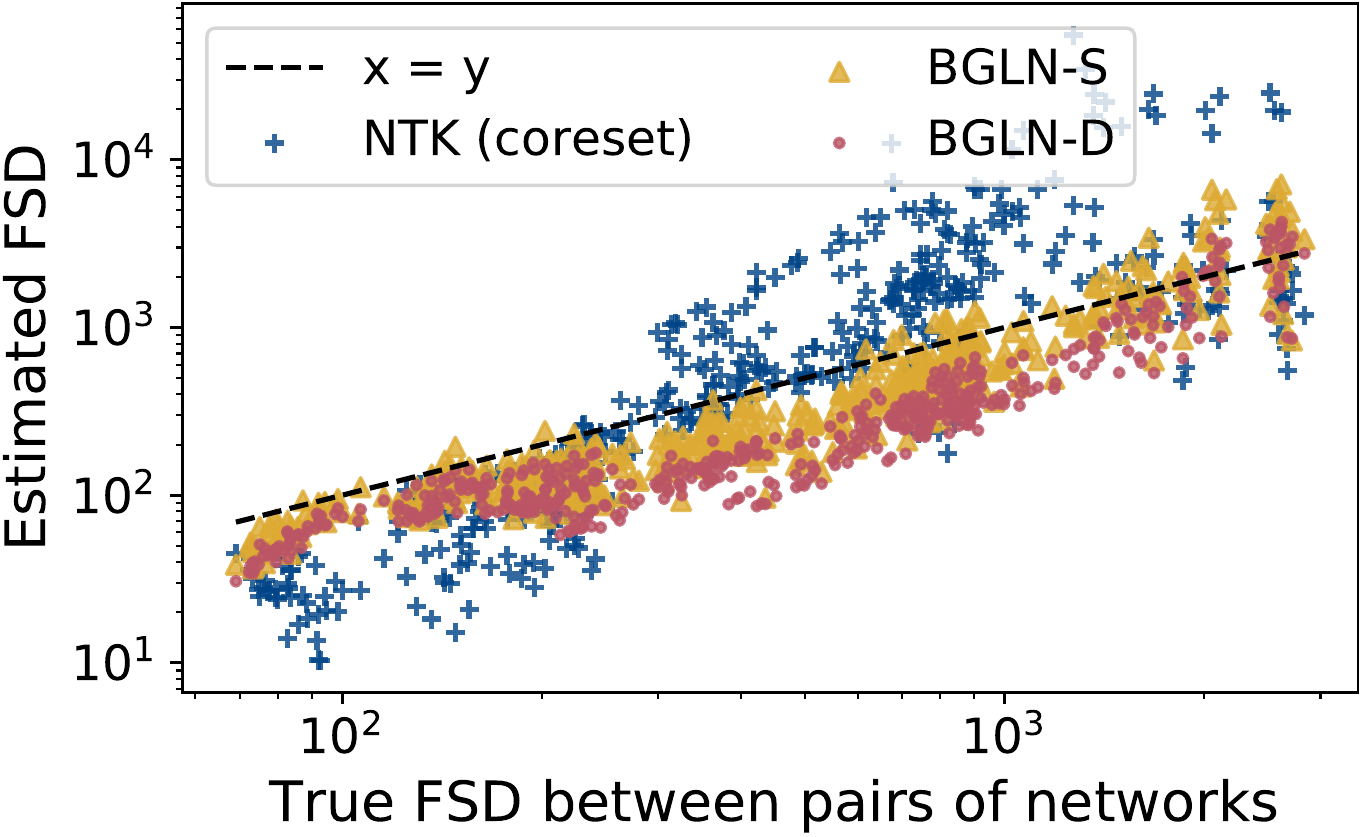}
    \vspace{-0.4cm}
\end{minipage}
\hfill
\begin{minipage}{0.32\textwidth}
    \small
    \centering
    \includegraphics[width=\linewidth]{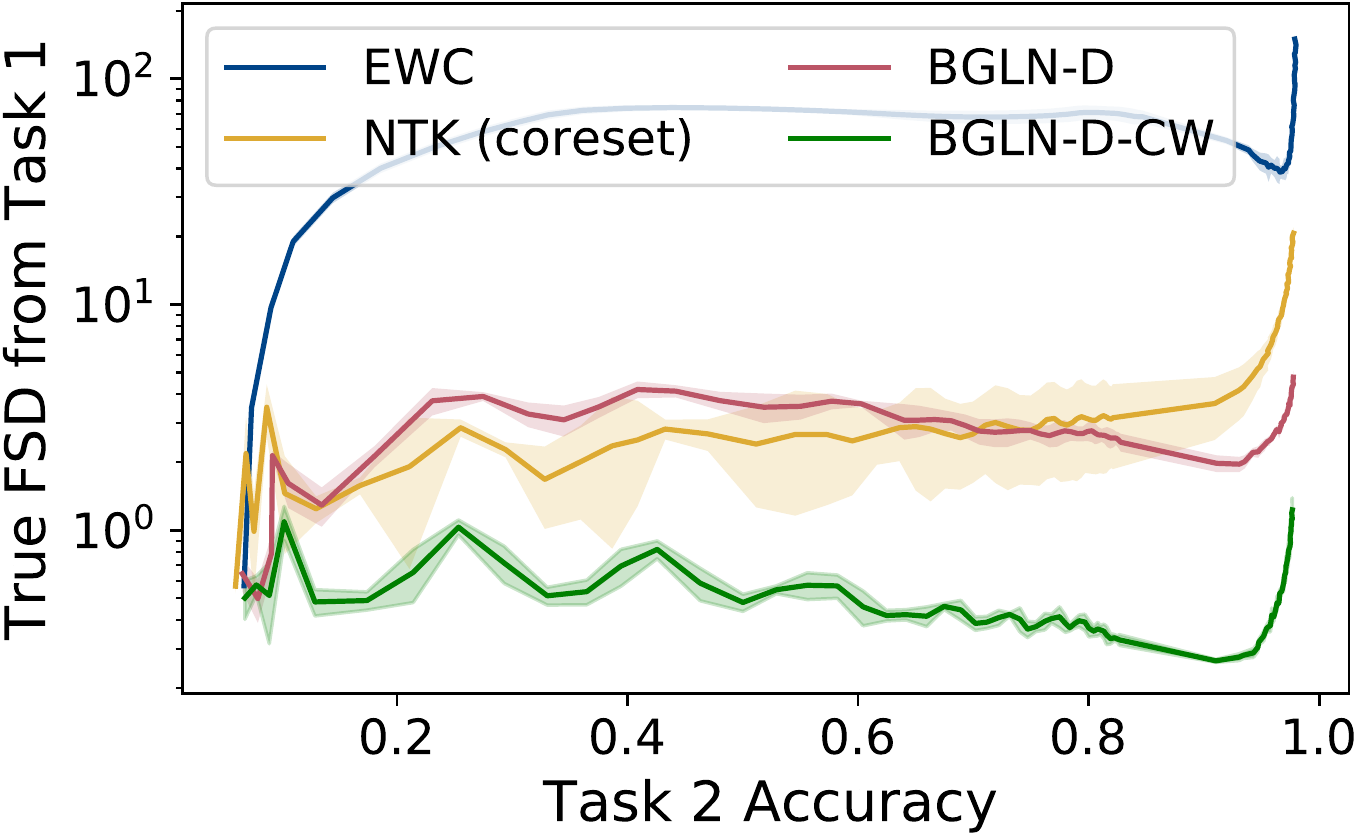}
    \vspace{-0.4cm}
\end{minipage}
\hfill 
\begin{minipage}{0.32\textwidth}
    \small
    \centering
    \includegraphics[width=\linewidth]{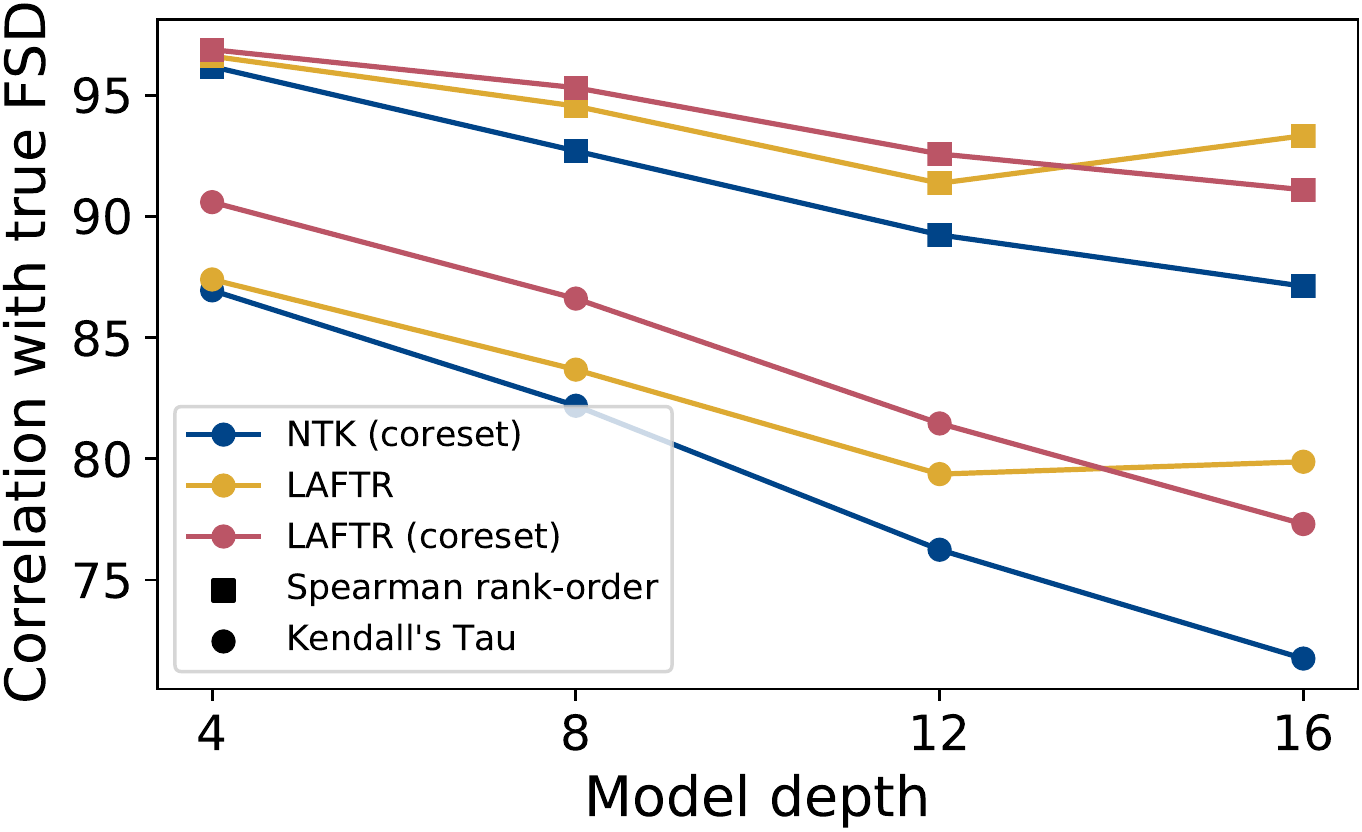}
    \vspace{-0.4cm}
\end{minipage}
\caption{Comparison of different FSD estimators. \textbf{(Left)} Compared to NTK, BGLN-S and BGLN-D consistently give closer FSD values to the true empirical FSD. \textbf{(Middle)} While training on task 2, FSD from the optimal task 1 parameters increases with task 2 accuracy. Optimizing BGLN-D and class-conditioned BGLN-D-CW effectively minimizes the true FSD. \textbf{(Right)} LAFTR has a higher correlation with true FSD than NTK, with a more significant advantage as network depth (and hence a number of nonlinear interactions) increases.}
\label{fig:est_comp}
\vspace{-0.2cm}
\end{figure*}

Figure \ref{fig:est_comp} (Left) shows that BGLN-S and BGLN-D consistently estimate the true FSD more accurately than NTK. Analogous analysis using CIFAR100 shows a similar trend in Figure \ref{fig:fsd_cifar_est}. We can also measure how the true FSD changes when different FSD estimates are optimized during training, as in Figure \ref{fig:est_comp} (Middle), where BGLN methods more effectively minimize true FSD as new task accuracy increases. Finally, we train networks of varying depths on CIFAR100 tasks and measure correlation (Spearman rank-order~\citep{spearman1961proof} and Kendall's Tau~\citep{kendall}) with true FSD. Figure \ref{fig:est_comp} (Right) shows that LAFTR has a higher correlation using both metrics, and its advantage over NTK increases with network depth, and hence with a number of nonlinear interactions between parameters. This corroborates our intuition that LAFTR captures nonlinearities that NTK is unable to account for.

\begin{table*}[!t]
\centering
\footnotesize
\begin{center}
\resizebox{\textwidth}{!}{
\begin{tabular}{lccccccc}
\toprule
\multicolumn{1}{l}{\bf Method} &\multicolumn{1}{c}{$\boldsymbol{\rho}$} &\multicolumn{1}{c}{\bf Coreset} &\multicolumn{1}{c}{\bf Bernoulli} &\multicolumn{1}{c}{\bf CW}  &\multicolumn{1}{c}{\bf Average Accuracy $\uparrow$} &\multicolumn{1}{c}{\bf Backward Transfer $\uparrow$} &\multicolumn{1}{c}{\bf Memory Cost} \\ 
\midrule 
\underline{Nonparametric} \\
\textbf{VCL (coreset)} & \multicolumn{4}{c}{$-$} & $67.40\pm0.60$ & $-$ & $2P + Nd$ \\
\textbf{VAR-GP (coreset)} & \multicolumn{4}{c}{$-$} & $-$ & $-$ & $2P+ Nd + C^2N^2$ \\
\textbf{FROMP (coreset)} & \multicolumn{4}{c}{$-$} & \third $76.20\pm0.20$ & \third $-2.60\pm0.90$ & $2P + Nd + C^2N^2$ \\
\textbf{S-FSVI (coreset)} & \multicolumn{4}{c}{$-$} & \second $77.60\pm0.20$ & \third $-2.50\pm0.20$ & $2P + Nd + C^2N^2$ \\
\textbf{NTK (coreset)} & KL & \multicolumn{3}{c}{$-$} & \second $77.61\pm0.20 $ & \third $-2.03\pm0.04$ & $2P + Nd$ \\
\textbf{LAFTR (coreset)} & KL & {\color{dkgreen} \cmark} & {\color{dkred} \xmark} & {\color{dkred} \xmark} & \first $\mathbf{78.33\pm0.01}$ & \first $\mathbf{-0.73\pm0.10}$ & $P + Nd$ \\
\textbf{BGLN-S (coreset)} & KL & {\color{dkgreen} \cmark} & {\color{dkgreen} \cmark} & {\color{dkred} \xmark} & $73.27\pm0.01$ & $-4.99\pm0.28$ & $P + A + Nd$ \\
\textbf{LAFTR (coreset)} & Euclidean & {\color{dkgreen} \cmark} & {\color{dkred} \xmark} & {\color{dkred} \xmark} & \third $76.22\pm0.01$ & \third $-2.64\pm0.40$ & $P + Nd$ \\ %

\midrule
\underline{Parametric} \\
\textbf{EWC} & \multicolumn{4}{c}{$-$} & $71.60\pm0.40$ & $-$ & $2P$\\
\textbf{OSLA} & \multicolumn{4}{c}{$-$} & $72.61$ & $-$ & $P + \sum_{l=1}^L p_l ^2$ \\
\textbf{VCL} &\multicolumn{4}{c}{$-$} & $-$ & $-$ & $2P$ \\ 
\textbf{LAFTR} & KL & {\color{dkred} \xmark} & {\color{dkred} \xmark} & {\color{dkred} \xmark} & \fourth $75.61\pm0.01$ & \second $-1.93\pm0.59$ & $P + d + d^2$ \\
\textbf{LAFTR-CW} & KL & {\color{dkred} \xmark} & {\color{dkred} \xmark} & {\color{dkgreen} \cmark} & \third $76.22\pm0.01$ & \second $-1.45\pm0.63$ & $P + C(d + d^2)$ \\
\textbf{BGLN-S} & KL & {\color{dkred} \xmark} & {\color{dkgreen} \cmark} & {\color{dkred} \xmark} & $72.37\pm0.01$ & $-8.20\pm0.04$ & $P + A + d + d^2$ \\
\textbf{BGLN-S-CW} & KL & {\color{dkred} \xmark} & {\color{dkgreen} \cmark} & {\color{dkgreen} \cmark} & $74.02\pm0.01$ & \third $-2.44\pm0.15$ & $P + C(A + d + d^2)$ \\
\textbf{LAFTR} & Euclidean & {\color{dkred} \xmark} & {\color{dkred} \xmark} & {\color{dkred} \xmark} & \fourth $75.51\pm0.01$ & \fourth $-3.12\pm0.44$ & $P + d + d^2$ \\
\textbf{BGLN-S} & Euclidean & {\color{dkred} \xmark} & {\color{dkgreen} \cmark} & {\color{dkred} \xmark} & $74.29\pm0.01$ & $-5.49\pm0.05$ & $P + A + d + d^2$ \\
\textbf{BGLN-S-CW} & Euclidean & {\color{dkred} \xmark} & {\color{dkgreen} \cmark} & {\color{dkgreen} \cmark} & \second $77.78\pm0.01$ & \second $-1.75\pm0.50$ & $P + C(A + d + d^2)$ \\
\hline 
\end{tabular}
}
\end{center}
\caption{Split CIFAR100: Average Accuracy and Backward Transfer. Notation for memory cost: $p_l$ = \# parameters in layer $l$, $P$ = \# parameters $= \sum_{l=1}^L p_l$, $A$ = \# activations $< P$, $d$ = data dimension, $N$ = coreset size, $C$ = \# classes.}
\label{table:kl_cifar}
\end{table*}

\subsection{Continual Learning}
\label{cl}
Recall the formulation of continual learning in terms of FSD, as described in equation~\ref{cl_eq}.
We visualize our method's comparative performance on 1-D regression with two sequential tasks shown in Figure \ref{fig:fsd_1d}. More realistically, we test our methods on standard benchmarks used in prior works~\citep{fromp, rudner2022continual}, with standard architectures for a fair comparison. See Appendix~\ref{app:cl_exp} for details on the datasets, architectures, and hyperparameters.
We evaluate average final accuracy across tasks, backward transfer~\citep{lopez2017gradient}
and memory cost.

\textbf{Toy Regression.} 
Figure \ref{fig:fsd_1d} shows the functions learned by different methods when sequentially trained on two one-dimensional regression tasks. LAFTR gives a better approximation of the learned function than NTK. When used to regularize the network, BGLN retains good predictions on both tasks, while EWC and exact parameter space linearization (NTK) suffer catastrophic forgetting. We hypothesize that this difference in performance is due to important nonlinearities between network parameters that EWC and NTK approximations are unable to capture. 

\textbf{Split and Permuted MNIST.} 
As shown in Tables \ref{table:avg_mnist} and \ref{table:bwt_mnist}, our LAFTR-based methods outperform other parametric methods (EWC, OSLA, and VCL) on Split and Permuted MNIST tasks and are competitive with the state-of-the-art (SOTA) nonparametric methods, in terms of average accuracy. Class-conditional approximations further boost performance and the diagonal approximation to input covariance (BGLN-S-Var, BGLN-D-Var) does not harm it significantly. With respect to the backward transfer, a more direct measure of forgetting, BGLN methods significantly outperform the SOTA. 
Finally, they are also amenable to successful adaptation to the nonparametric setting when a coreset is available for use in place of Gaussian samples.

\textbf{Split CIFAR100.}
We consider the much more challenging Split CIFAR100 task to compare our method to existing approaches and tease apart the effects of our algorithmic choices and approximations. 
Table \ref{table:kl_cifar} summarizes these results and analytically compares the memory costs associated with each method (see Appendix \ref{app:mem} for details). \textbf{Coreset} refers to a coreset of real inputs vs. Gaussian samples, \textbf{Bernoulli} refers to Bernoulli activations vs. simply passing preactivations through the ReLU function (termed LAFTR here) and \textbf{CW} refers to the class-conditonal estimate. Our empirical analysis shows that given a random coreset of the same size as comparable methods, LAFTR outperforms the SOTA as well as the NTK baseline on average accuracy and backward transfer significantly. Other LAFTR and BGLN variants remain competitive with prior methods while incurring lower memory costs. We also observe that performance is hurt to some extent by the Gaussian and Bernoulli modeling assumptions, while it is improved by class-conditioning. We present a more fine-grained task-wise comparison of accuracies in Figure \ref{app:taskwise_fig} and results on a longer task sequence in Table \ref{table:task_seq} of Appendix \ref{app:taskwise}.

\textbf{Memory Gains.}
To demonstrate the memory gains for our LAFTR and LAFTR (coreset) methods in practical scenarios, we compute the percentage reduction in memory costs relative to those of typical nonparametric methods. Specifically, we fix the network architecture to be a convolutional network used for training Split CIFAR100, and select reasonable values of coreset size, number of classes per task and data dimension from a range encountered in practice. We then use equations in Table \ref{table:kl_cifar} to compute $100 \times \frac{B - L}{B}$, where $B$ and $L$ are the memory costs of a nonparametric method and a LAFTR variant, respectively. As summarized in Table \ref{memory_gain}, the gains in memory complexity enjoyed by LAFTR methods increase as coreset size is increased, number of classes are increased or data dimension is decreased. Further results on Split CIFAR100 performance as amount of stored information (for example, coreset size) is varied are included in Appendix \ref{mem_abl}.

\begin{table*}[t]
\centering
\scriptsize
\begin{subtable}{}
    \resizebox{0.80\columnwidth}{!}{
    \begin{tabular}{c|ccc|c}
    \toprule
    \diagbox{\textbf{C}}{\textbf{d}}  & \textbf{1000} & \textbf{2000} & \textbf{3000} & \textbf{N}   \\ \midrule
    \multirow{2}{*}{\textbf{10}} & $66.27$         & $24.08$         & $-43.71$        & \textbf{200} \\
                                 & $74.84$         & $43.26$         & $-7.60$         & \textbf{250} \\ \midrule
    \multirow{2}{*}{\textbf{20}} & $87.86$         & $72.15$         & $46.33$         & \textbf{200} \\
                                 & $91.81$         & $81.18$         & $63.68$         & \textbf{250} \\ \midrule
    \multirow{2}{*}{\textbf{50}} & $97.78$         & $94.87$         & $90.03$         & \textbf{200} \\
                                 & $98.57$         & $96.69$         & $93.56$         & \textbf{250} \\ \bottomrule
    \end{tabular}
    }
\end{subtable}
\begin{subtable}{}
    \resizebox{0.77\columnwidth}{!}{
    \begin{tabular}{c|ccc|c}
    \toprule
    \diagbox{\textbf{C}}{\textbf{d}}  & \textbf{1000} & \textbf{2000} & \textbf{3000} & \textbf{N}   \\ \midrule
    \multirow{2}{*}{\textbf{10}} & $78.14$         & $75.89$         & $73.77$        & \textbf{200} \\
                                 & $83.14$         & $80.90$         & $78.79$         & \textbf{250} \\ \midrule
    \multirow{2}{*}{\textbf{20}} & $92.13$         & $91.15$         & $90.20$         & \textbf{200} \\
                                 & $94.51$         & $93.67$         & $92.84$         & \textbf{250} \\ \midrule
    \multirow{2}{*}{\textbf{50}} & $98.56$         & $98.37$         & $98.18$         & \textbf{200} \\
                                 & $99.04$         & $98.88$         & $98.73$         & \textbf{250} \\ \bottomrule
    \end{tabular}
    }
\end{subtable}
\caption{Percentage reduction in memory cost of LAFTR (\textbf{Left}) and LAFTR (coreset) (\textbf{Right}) relative to typical nonparametric methods for a fixed network architecture and varying values of $C$ (number of classes), $d$ (data dimension) and $N$ (coreset size). Higher percentages are better. In both cases, gains in memory complexity increase as $N$ increases, $C$ increases or $d$ decreases.}
\label{memory_gain}
\end{table*}

\subsection{Influence Function Estimation}
\label{inf_exp}

To further assess BGLN's applicability to other settings involving FSD estimation and regularization, we consider influence function estimation~\citep{cook1979influential,hampel, koh2017understanding}. 
Given parameters $\rvtheta_0$ trained on dataset $\mathcal{D}_{\text{train}}$ of size $N$,
influence functions approximate the parameters $\rvtheta_-$ that would be obtained by training without a particular point $(\vx, \vy) \in \mathcal{D}_{\text{train}}$. The difference in loss between $\rvtheta_0$ and $\rvtheta_{-}$ is an indicator of the influence of $(\vx, \vy)$ on the trained network.

\citet{bae2022influence} show that influence functions in neural networks can be formulated as solving for the proximal Bregman response function (PBRF):
\vspace{-0.2cm}
\begin{align}  
\label{eq:pbrf}
    \nonumber
    \rvtheta_{-} = \argmin_{\rvtheta \in \mathbb{R}^d} &
     - \frac{1}{N} \mathcal{L}(f(\vx, \rvtheta), \vy) +
    D_B (\rvtheta, \rvtheta_0, p_{\text{train}})\\ 
    &\quad+ \frac{\lambda}{2} \|\rvtheta - \rvtheta_0 \|^2 .
\end{align}
Here, the first term maximizes the loss of the data point we are interested in removing. The second term is the Bregman divergence defined on network outputs and measures the FSD between $\rvtheta$ and $\rvtheta_0$ over training distribution $p_{\text{train}}$ similar to the FSD term 
as defined in equation~\ref{fsd_def}. 
For standard loss functions like squared error and cross-entropy, the Bregman divergence term is equivalent to the soft training error where the original targets are replaced with soft targets produced by $\rvtheta_0$. Finally, the last term is a proximity term with strength $\lambda > 0$, which prevents large changes in weight space. 
Intuitively, the PBRF maximizes the loss of data we would like to remove while constraining the network in both function and weight space so that the predictions and losses of other training examples remain unaffected.

\begin{table}[t]
    \centering
    \scriptsize
    \begin{adjustbox}{width=0.85\columnwidth}
    \begin{tabular}{@{}ccccccc@{}}
    \toprule
    \textbf{Dataset} & \multicolumn{2}{c}{\textbf{EWC}} & \multicolumn{2}{c}{\textbf{CG}} & \multicolumn{2}{c}{\textbf{BGLN-D}} \\ \cmidrule(l){2-7} 
     & P & S & P & S & P & S \\ \midrule
    Concrete & 0.78 & 0.57 & 0.92 & 0.94 & \textbf{0.96}  &  \textbf{0.97}\\ 
    Energy & 0.68 & 0.39 & 0.97  & 0.98  & \textbf{0.99}  & \textbf{0.98}  \\ 
    Housing & 0.86 & 0.33 & 0.92 & \textbf{0.89} & \textbf{0.95}  & 0.83\\
    Kinetics & 0.36 & 0.30 & 0.88 & 0.86 & \textbf{0.99}  & \textbf{0.99}\\ 
    Wine & 0.97 & 0.70 & 0.99 & \textbf{0.94} & \textbf{0.99}  & 0.90\\ 
    \bottomrule
    \end{tabular}
    \end{adjustbox}
    \caption{Comparison of training loss differences computed by EWC, CG and BGLN-D. We show Pearson (P) and Spearman rank-order (S) correlations with the PBRF estimates.}
    \vspace{-0.2cm}
    \label{table:corr_inf}
\end{table}
Existing approaches for this optimization face two key challenges: (1) the entire training dataset must be stored and iterated over, requiring as many forward passes as there are mini-batches and (2) techniques like nonlinear Conjugate Gradient (CG)~\citep{hager2006survey} do not work well with the stochastic gradients produced by sampling batches of data. LAFTR enables estimating the PBRF (or FSD) by storing only the first two data moments, requires just a single forward pass to compute it and provides a deterministic function to optimize, implemented as BGLN-D.

\begin{figure}[t]
    \centering
    \includegraphics[width=0.95\columnwidth]{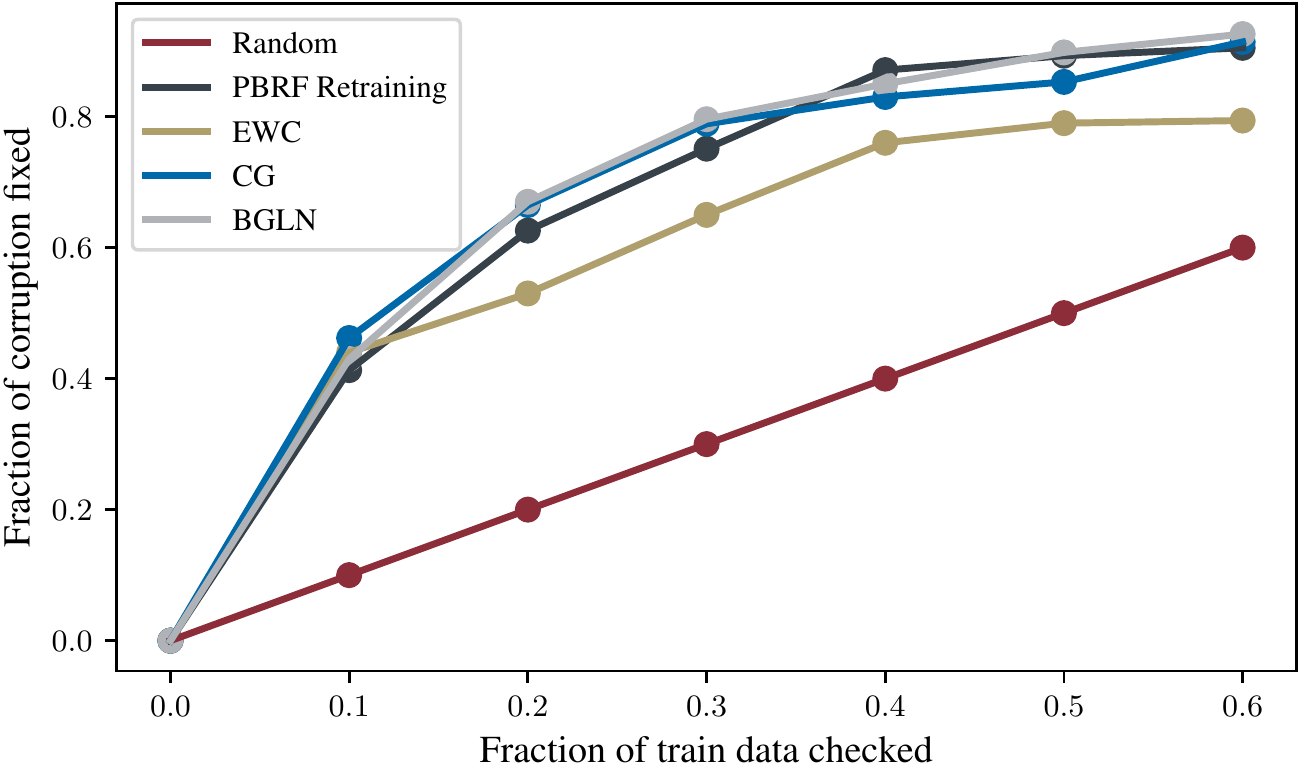}
    \vspace{-0.3cm}
    \caption{Effectiveness of BGLN in detecting mislabeled examples. BGLN can approximate the FSD term in the PBRF objective accurately and be used in applications involving influence functions without explicitly storing or iterating over the dataset. }
    \vspace{-0.2cm}
    \label{fig:inf_det}
\end{figure}

\textbf{Regression.} We first train a MLP with two hidden layers and ReLU activations for 200 epochs on regression datasets from the UCI benchmark~\citep{Dua:2019}. Then, we randomly select 50 independent data points to be removed. For each removed point, we sample batches and use a Stochastic Gradient Descent (SGD) optimizer to minimize the PBRF objective and compute the difference in loss after removing that data point, commonly referred to as the self-influence score~\citep{koh2017understanding,schioppa2022scaling}. 
Next, we follow the same procedure as above but approximate the FSD term in the PBRF objective with EWC, CG~\citep{koh2017understanding} and BGLN-D. Since the direct minimization of PBRF can be considered as the ground truth for influence estimation, we compare the alignment of these methods' estimates with that of 
PBRF via Pearson correlation~\citep{sedgwick2012pearson} and Spearman rank-order correlation~\citep{spearman1961proof}. The results are shown in Table~\ref{table:corr_inf}. 
Without having to iterate over or store the entire dataset, BGLN-D correlates with PBRF more strongly than EWC and CG~\citep{koh2017understanding}, which can be seen as minimizing a linearized version of the PBRF objective.

\textbf{Mislabeled Example Detection.}
Influence function estimators are commonly evaluated in terms of their ability to identify mislabeled examples. Intuitively, if some fraction of the training labels is corrupted, they would behave as outliers and have a more significant influence on the training loss (self-influence score). One approach to efficiently detect and correct these examples is to prioritize and examine training inputs with higher self-influence scores. 
Following the evaluation setup from~\citet{bae2022influence}, we use 10\% of the MNIST dataset and corrupt 10\% of it by assigning random labels to it. We train a two layer MLP with 1024 hidden units and ReLU activations using SGD with a batch size of 128. Then, we use EWC, CG and BGLN-D to approximate the FSD term in equation~\ref{eq:pbrf} and compute individual self-influence scores. We also compare these methods against a baseline of randomly sampling data points to check for corruption. The results are summarized in Figure~\ref{fig:inf_det}. BGLN-D significantly outperforms the random baseline and EWC and closely matches the oracle PBRF and CG, while being much faster, cheaper and more memory-efficient.

\section{Conclusions}
\label{concl}

In this work, we addressed the problem of compactly summarizing a model's predictions on a given dataset, and formulated it as approximating neural network FSD. We developed the Linearized Activation Function TRick as an improvement over network linearization in parameter space and proposed novel parametric methods, BGLN, to estimate FSD. Our methods capture nonlinearities between network parameters, are much more memory-efficient than prior works and are amenable to adaptation to the nonparametric setting when a coreset of data is available. 

We empirically show that LAFTR-based estimates are highly correlated with the true FSD across several settings. In continual learning, our methods outcompete existing methods without storing any data samples. Further, in influence function estimation, they estimate influence-scores with high correlation and can efficiently detect mislabeled examples without expensive iteration over the whole dataset. 

Extending the formulation of FSD approximation to other applications like model editing or unlearning are exciting research avenues. 
We hope that our work inspires methods to further enhance memory and computational efficiency in settings where estimating or constraining FSD is relevant.

\section*{Acknowledgements}
We would like to thank Florian Shkurti for useful discussions, Cem Anil for feedback on the draft, and Gerald Shen for assistance with the compute environment. Resources used in preparing this research were provided, in part, by the Province of Ontario, the Government of Canada through CIFAR, and companies sponsoring the Vector Institute (\url{www.vectorinstitute.ai/partners}).

\bibliography{references}
\bibliographystyle{icml2023}

\newpage
\appendix
\onecolumn
\section*{Appendix}

\section{Notation}

\bgroup
\def\arraystretch{1.5}
\begin{tabular}{p{1in}p{5.25in}}
$\displaystyle f$ & Function corresponding to a neural network\\
$\displaystyle \phi$ & ReLU activation function\\
$\displaystyle \phi'$ & Derivative of a function $\phi$\\
$\displaystyle \vs_l$ & Preactivations at layer $l$\\
$\displaystyle \va_l = \phi(\vs_l)$ & Activations at layer $l$\\
$\displaystyle \rvx$ & Input data\\
$\displaystyle \vy$ & Target data\\
$\displaystyle \rvx^{(i)}$ & Input data point $i$\\
$\displaystyle p_{\text{data}}$ & Data distribution from which $\rvx$ is sampled\\
$\displaystyle d$ & Data dimension\\
$\displaystyle N$ & Number of data points in a coreset\\
$\displaystyle L$ & Number of layers in a network\\
$\displaystyle C$ & Number of classes in a classification task\\
$\displaystyle T$ & Number of tasks\\
$\displaystyle \rvtheta$ & Parameters of a neural network\\
$\displaystyle \rvtheta_t$ & Parameters obtained after training on task $t$\\
$\displaystyle p_l$ & Number of parameters in layer $l$\\
$\displaystyle P$ & Total number of parameters in a network $=\sum_{l=1}^L p_l$\\
$\displaystyle \vz = f(\rvx; \rvtheta)$ & Prediction of the network $f$ on $\rvx$ parameterized by $\rvtheta$ \\
$\displaystyle \rho$ & Output space distance, for instance Euclidean distance\\
$\displaystyle D(\rvtheta_0, \rvtheta_1, p_{\text{data}})$ & FSD between networks parameterized by $\rvtheta_0$ and $\rvtheta_1$ over data distribution $p_{\text{data}}$\\
$\displaystyle \mG_{\rvtheta}$ & Weight space metric matrix\\
$\displaystyle \mF_{\rvtheta}$ & Fisher information matrix\\
$\displaystyle \rvm$ & Mask vector or Bernoulli random variable\\
$\displaystyle \boldsymbol{\mu}$ & Bernoulli mean parameter\\

\end{tabular}
\egroup

\section{Recursion Equations for BGLN-D}
\label{app:bgln-d}
We derive the deterministic version of our algorithm by taking expectations and covariances for the quantities in equations \ref{s0} to \ref{delta_a} (rewritten using Bernoulli variables). We use linearity of expectations and our Bernoulli modeling approximation. We also assume $\mathrm{Cov}(\va_0, \Delta \va)$ is close to 0 and ignore it in our computations. This assumption is tested empirically in our experiments and we find that it does not severely move the FSD estimate away from the true empirical FSD (see Figures \ref{fig:est_comp} and \ref{fig:fsd_cifar_est}).
The complete steps for BGLN-D computations are given in Algorithm \ref{determ_alg}.

\begin{algorithm}[tbh]
    \caption{BGLN-D}
    \label{determ_alg}
    \begin{algorithmic}
        \REQUIRE {$\mathbb{E}[\rvx], \mathrm{Cov}(\rvx), \{\mW, \vb\}_1^{L}, \{\boldsymbol{\mu}\}_1^{L-1}$}
        \STATE $\mathbb{E} [\va_0], \mathbb{E} [\va_1] \gets \mathbb{E}[\rvx]$
        \STATE {$\mathrm{Cov} (\va_0), \mathrm{Cov} (\va_1) \gets \mathrm{Cov}(\rvx)$} 
        \STATE {$\mathbb{E}[\Delta \va], \mathrm{Cov} (\Delta \va) \gets 0$}
        \FOR{$l \gets 1$ to $L-1$}  
            \STATE {$\mathbb{E} [\vs_0] \gets \mW_0^{(l)} \mathbb{E} [\va_0] + \vb_0^{(l)}$} 
            \STATE {$\mathbb{E} [\Delta \vs] \gets \Delta \mW^{(l)} \mathbb{E} [\va_0] + \mW_1^{(l)} \mathbb{E}[\Delta \va] + \Delta \vb^{(l)}$}
            \STATE {$\mathbb{E} [\va_0] \gets \boldsymbol{\mu}^{(l)} \odot \mathbb{E} [\vs_0]$}  
            \STATE {$\mathbb{E} [\Delta \va] \gets \boldsymbol{\mu}^{(l)} \odot \mathbb{E} [\Delta \vs]$}
            \STATE {$\mathrm{Cov} (\vs_0) \gets \mW_0^{(l)} \mathrm{Cov} (\va_0) \mW_0^{(l)T}$} 
            \STATE {$\mathrm{Cov} (\Delta \vs) \gets \Delta \mW^{(l)} \mathrm{Cov}(\va_0) \Delta \mW^{(l)T} + \mW_1^{(l)} \mathrm{Cov} (\Delta \va) \mW_1^{(l)T}$} 
            \STATE {$\mathrm{Cov} (\va_0) \gets (\boldsymbol{\mu}^{(l)} \boldsymbol{\mu}^{(l)T}) \odot \mathrm{Cov} (\vs_0)$} 
            \STATE {$\mathrm{Cov} (\Delta \va) \gets (\boldsymbol{\mu}^{(l)} \boldsymbol{\mu}^{(l)T}) \odot \mathrm{Cov} (\Delta \vs)$}
        \ENDFOR
        \STATE {$\mathbb{E} [\Delta \vz] \gets \Delta \mW^{(L)} \mathbb{E} [\va_0] + \mW_1^{(L)} \mathbb{E}[\Delta \va] + \Delta \vb^{(L)}$}
        \STATE {$\mathrm{Cov} (\Delta \vz) \gets \Delta \mW^{(L)} \mathrm{Cov}(\va_0) \Delta \mW^{(L)T} + \mW_1^{(L)} \mathrm{Cov} (\Delta \va) \mW_1^{(L)T}$} 
        \STATE \textbf{return } {$\frac{1}{2} ||\mathbb{E} [\Delta \vz]||^2 + \frac{1}{2} \textit{tr } (\mathrm{Cov} (\Delta \vz))$}
    \end{algorithmic}
\end{algorithm}

\begin{algorithm}[tbh]
    \caption{BGLN-S (Conv)}
    \label{conv_alg}
    \begin{algorithmic}[1]
        \REQUIRE {$\mathbb{E}[\rvx], \mathrm{Cov}(\rvx), \{\texttt{layer}\}_1^{L-1}, \{\mu\}_1^{L-1}$}
        \STATE {$\mathbb{E} [\va_0], \mathbb{E} [\va_1] \gets \mathbb{E}[\rvx]$} 
        \STATE {$\mathrm{Cov} (\va_0), \mathrm{Cov} (\va_1) \gets \mathrm{Cov}(\rvx)$} 
        \STATE {$\va_0, \va_1 \sim \mathcal{N}(\mathbb{E}[\va_0], \mathrm{Cov}(\va_0))$} 
        \STATE {$\Delta a \gets 0$}
        \FOR{$l \gets 1$ to $L-1$}  
            \IF {$\texttt{layer}$ is $\texttt{Conv}$ or $\texttt{FC}$}
                \STATE {$\vs_0 \gets \texttt{layer}(\va_0, \texttt{grad=False})$}
                \STATE {$\vs_1 \gets \texttt{layer}(\va_1)$}
            \ELSIF {$\texttt{layer}$ is $\texttt{ReLU}$}
                \STATE {$\Delta \vs = \vs_1 - \vs_0$}
                \STATE {$\rvm \sim Ber(\mu^{(l)})$}
                \STATE {$\Delta \va \gets \rvm \odot \Delta \vs$}
            \ELSE 
                \STATE {$\va_0 \gets \texttt{layer}(\va_0)$}
                \STATE {$\va_1 \gets \texttt{layer}(\va_1)$}
            \ENDIF
        \ENDFOR
        \STATE {$\Delta \vz \gets \vs_1 - \vs_0$} 
        \STATE \textbf{return } {$\frac{1}{2}||\Delta \vz||^2$}
    \end{algorithmic}
\end{algorithm}

\section{Generalization to Convolutional Networks}
\label{app:conv}

The generalization of BGLN-S to convolutional networks involves passing the inputs sampled using the data moments through the network. In convolutional networks, ReLU activation is usually applied after the convolutional and the fully connected layers. At each ReLU step, we use the Bernoulli mean parameters to sample activation signs and obtain the difference of activations, $\Delta \va$. 
Finally, the Euclidean distance (or KL divergence) between the final layer outputs leads to the stochastic estimate of the FSD. The complete procedure is shown in Algorithm~\ref{conv_alg}.

\section{Comparing FSD Estimators}

\begin{figure}[!t]
\small
\centering
\includegraphics[width=0.75\linewidth]{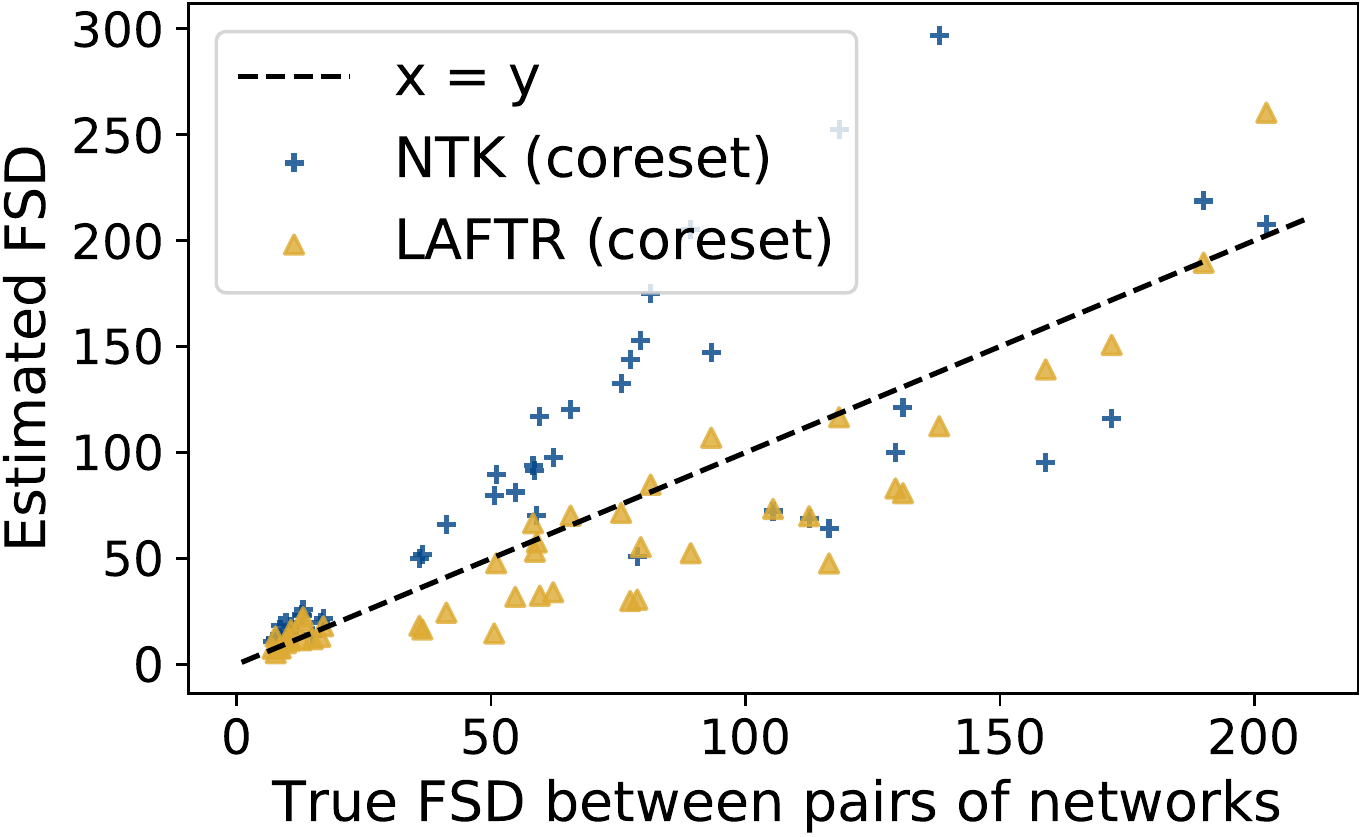}
\caption{Estimating true FSD with LAFTR and NTK, given the same coreset of inputs, on networks trained using Split CIFAR100.}
\label{fig:fsd_cifar_est}
\end{figure}

To directly compare the NTK approximation with the linearized activation function trick, we compute the estimated FSD between pairs of networks using these two kinds of linearization, when provided with the same information, and plot them against the true empirical FSD. Hence, both estimators are provided with the same coreset of datapoints. In the case of BGLN-S (coreset), activations are computed using this coreset directly instead of Bernoulli sampling. Figure \ref{fig:fsd_cifar_est} visualizes this comparison for networks trained using tasks in Split CIFAR100. BGLN-S estimates correlate better with the true FSD than NTK, hence corroborating our intuition about linearizing activation functions. To quantify this difference, we also measure the Spearman rank-order and Kendall's Tau correlation coefficients for each estimator with the true FSD. BGLN-S obtains values 96.36 and 79.19, respectively, outperforming NTK, which obtains 86.42 and 71.71, respectively.

\section{Continual Learning}

\subsection{Memory Cost Analysis}
\label{app:mem}

\begin{wraptable}[8]{r}{0.3\textwidth}
    \vspace{-0.6cm}
    \centering
    \scriptsize
    \caption{Memory cost notation.}
    \begin{adjustbox}{width=0.3\textwidth}
    \begin{tabular}{lc}
    \toprule
    $p_l$ & \# parameters in layer $l$\\
    $P$ & \# parameters $= \sum_{l=1}^L p_l$ \\
    $A$ & \# activations $< P$ \\
    $d$ & data dimension \\
    $N$ & coreset size \\
    $C$ & \# classes \\
    \hline
    \end{tabular}
    \end{adjustbox}
    \label{table:mem}
    \vspace{-0.6cm}
\end{wraptable}

We follow the notation in Table ~\ref{table:mem} to denote $p_l$ as the number of parameters in layer $l$, $P=\sum_{l=1}^L p_l$ as the total number of parameters in the network, $A$ as the number of activations in the network (note that $A < P$), $d$ as the data dimension, $N$ as the number of samples in a coreset, and $C$ as the number of classes in the continual learning classification setting. We can now write analytic expressions for the memory cost incurred by the different methods considered in the continual learning experiments, as shown in Table \ref{table:kl_cifar}. Below we arrive at these expressions for each task that the model is continually trained on.
\begin{itemize}
    \item \textbf{EWC:} EWC requires storing one value for each parameter of the network and one value for each diagonal element of the $P \times P$ Fisher information matrix, resulting in a cost of $2P$.
    \item \textbf{OSLA:} OSLA approximates the Fisher information matrix as a block diagonal matrix, storing $p_l^2$ elements for each block corresponding to layer $l$. This gives a cost of $P + \sum_{l=1}^L p_l ^2$. 
    \item \textbf{VCL:} VCL stores two pieces of information for the variational distribution for each parameter, one for the mean and one for the variance in the diagonal approximation, incurring a cost equal to $2P$.
    \item \textbf{VCL (coreset):} The ``coreset" variant of VCL additionally stores $N$ datapoints, increasing the memory cost by $Nd$.
    \item \textbf{VAR-GP, FROMP, S-FSVI:} These methods all store two values per parameter, similar to VCL. Further, they require a coreset of datapoints and/or inducing points, as well as a $NC \times NC$ kernel matrix. 
    \item \textbf{NTK (coreset):} The NTK approximation stores one value for each parameter of the network and one for the $P-$dimensional Jacobian-vector product used to linearize the network. It further requires a coreset of $N$ datapoints, result in a $2P + Nd$ cost.
    \item \textbf{LAFTR:} LAFTR methods store each parameter once and the first two moments of the data, incurring a cost of $P + d + d^2$.
    \item \textbf{LAFTR-CW:} The classwise variants of LAFTR store separate data moments for each class, which scales those corresponding memory costs by C, i.e., $P + C(d + d^2)$.
    \item \textbf{LAFTR (coreset):} The ``coreset"" ablation of LAFTR requires storing $N$ datapoints instead of the data moments, giving a $P + Nd$ cost.
    \item \textbf{BGLN-S, BGLN-D:} BGLN methods store each parameter once, a Bernoulli mean value for each activation and the first two moments of the data, incurring a cost of $P + A + d + d^2$.
    \item \textbf{BGLN-S-CW, BGLN-D-CW:} The classwise variants of our methods store separate Bernoulli means and data moments for each class, which scales those corresponding memory costs by C, i.e., $P + C(A + d + d^2)$.
    \item \textbf{BGLN-S-Var, BGLN-D-Var:} The ``Var" variants of our methods make a diagonal approximation to the data covariance (second moment), hence storing only $d$ values for it. This further reduces memory cost to $P + A 2d$.
    \item \textbf{BGLN-S (coreset):} The ``coreset'' ablation of BGLN-S requires storing $N$ datapoints instead of the data moments, giving a $P + A + Nd$ cost.
\end{itemize}

\subsection{Experimental Details}
\label{app:cl_exp}
\textbf{Datasets.} Split MNIST consists of five binary prediction tasks to classify non-overlapping pairs of MNIST digits~\citep{deng2012mnist}. Permuted MNIST is a sequence of ten tasks to classify ten digits, with a different fixed random permutation applied to the pixels of all training images for each task. Finally, Split CIFAR100 consists of six ten-way classification tasks, with the first being CIFAR10~\citep{cifar10}, and subsequent ones containing ten non-overlapping classes each from the CIFAR100 dataset~\citep{cifar100}.  

\textbf{Architectures.} We use standard architectures used by existing methods for fair comparison. For regression and the MNIST experiments, we use a MLP with two fully connected layers and ReLU activation. For Split CIFAR100, we use a network with four convolutional layers, followed by two fully connected layers, and ReLU activation after each. For MNIST tasks, we compute the FSD with Euclidean output space metric between logits. For CIFAR100, we do the same between softmax outputs. Both Split MNIST and Split CIFAR100 models have a multiheaded final layer. Note that the VAR-GP method included in the results in Section \ref{expts} is specific only to singleheaded architectures and does not generalize to the multiheaded networks that our methods and all other comparable methods use.

\textbf{Hyperparameters.} We have performed a grid search over some key hyperparameters and used the ones that resulted in the best final average accuracy across all tasks. All hyperparameter search was done with random seed 42. We then took that best set of hyperparameters, repeated our experiments on seeds 20, 21, 22, and reported the average and standard deviation of our results. 

For all nonparametric methods that store and use a coreset of datapoints, we use 40 points for MNIST datasets and 200 points for CIFAR 100, in accordance with standard protocol followed by comparable methods.

For the learning rate, we used $0.001$ for all CL experiments except the BGLN-S method for Split MNIST and BGLN-D method for Permuted MNIST, where we used $0.0001$ instead. 

We used the same number of epochs on each CL task and the exact numbers are reported in Table~\ref{table:epochs}. On the first task, all MNIST experiments used the same number of epochs as the subsequent CL tasks while CIFAR100 experiments used 200 epochs on the first task.

To compute the Bernoulli mean parameters for our stochastic gating implementation, we used simple averaging as the default, but also explored exponential moving averaging. While there was not much difference in performance, we report the momentum values that reproduce our results.  All MNIST experiments had a momentum value of $1/\texttt{batch\_size}$. Note that this momentum value of $1/\texttt{batch\_size}$ corresponds to simple moving average. For CIFAR100 experiments, we used $0.99$ for BGLN-S (CW) and NTK, and $1/\texttt{batch\_size}$ for BGLN-S. 

For each method and dataset, the scaling factor for FSD penalty, $\lambda_{\text{FSD}}$, is reported in Table~\ref{table:fsd_scale}. Similarly, batch size is reported in Table~\ref{table:bs}.

\begin{table}[!t]
\caption{CL tasks training epochs used in CL experiments.}
\begin{center}
\begin{tabular}{lccc}
\toprule
\multicolumn{1}{l}{\bf Method}  &\multicolumn{1}{c}{\bf Split MNIST} &\multicolumn{1}{c}{\bf Permuted MNIST} &\multicolumn{1}{c}{\bf Split CIFAR100} \\ 
\hline \\

\textbf{NTK (coreset)} & $15$ & $5$ & $80$ \\
\textbf{BGLN-S} & $15$ & $15$ & $80$ \\
\textbf{BGLN-D} & $15$ & $15$ & - \\
\textbf{BGLN-S-CW} & $15$ & $15$ & $50$ \\
\textbf{BGLN-D-CW} & $15$ & $15$ & - \\
\hline
\end{tabular}
\vspace{-0.2cm}
\end{center}
\label{table:epochs}
\end{table}

\begin{table}[!t]
\caption{FSD scale used in CL experiments.}
\begin{center}
\begin{tabular}{lccc}
\toprule
\multicolumn{1}{l}{\bf Method}  &\multicolumn{1}{c}{\bf Split MNIST} &\multicolumn{1}{c}{\bf Permuted MNIST} &\multicolumn{1}{c}{\bf Split CIFAR100} \\ 
\hline \\

\textbf{NTK (coreset)} & $1$ & $1$ & $0.005$ \\
\textbf{BGLN-S} & $5$ & $1$ & $1$ \\
\textbf{BGLN-D} & $0.1$ & $0.005$ & - \\
\textbf{BGLN-S-CW} & $2$ & $1$ & $10$ \\
\textbf{BGLN-D-CW} & $0.1$ & $0.005$ & - \\
\hline
\end{tabular}
\vspace{-0.2cm}
\end{center}
\label{table:fsd_scale}
\end{table}

\begin{table}[!t]
\caption{Batch size used in CL experiments.}
\begin{center}
\begin{tabular}{lccc}
\toprule
\multicolumn{1}{l}{\bf Method}  &\multicolumn{1}{c}{\bf Split MNIST} &\multicolumn{1}{c}{\bf Permuted MNIST} &\multicolumn{1}{c}{\bf Split CIFAR100} \\ 
\hline \\

\textbf{NTK (coreset)} & $256$ & $256 $ & $512$ \\
\textbf{BGLN-S} & $32$ & $128$ & $512$ \\
\textbf{BGLN-D} & $32$ & $128$ & - \\
\textbf{BGLN-S-CW} & $32$ & $128$ & $512$ \\
\textbf{BGLN-D-CW} & $32$ & $128$ & - \\
\hline
\end{tabular}
\vspace{-0.2cm}
\end{center}
\label{table:bs}
\end{table}
 
\textbf{Evaluation Metrics.} In addition to average accuracy over tasks, we measure the backward transfer metric. For $T$ tasks, let $R_{i, j}$ be the classification accuracy on task $t_j$ after training on task $t_i$. Then, backward transfer is given by the following formula.
\[\frac{1}{T-1} \sum_{i=1}^{T-1} R_{T,i} - R_{i,i}\]

\subsection{Task-wise classification and task sequence length}
\label{app:taskwise}

\begin{figure}[!t]
    \centering
    \includegraphics[width=0.8\textwidth]{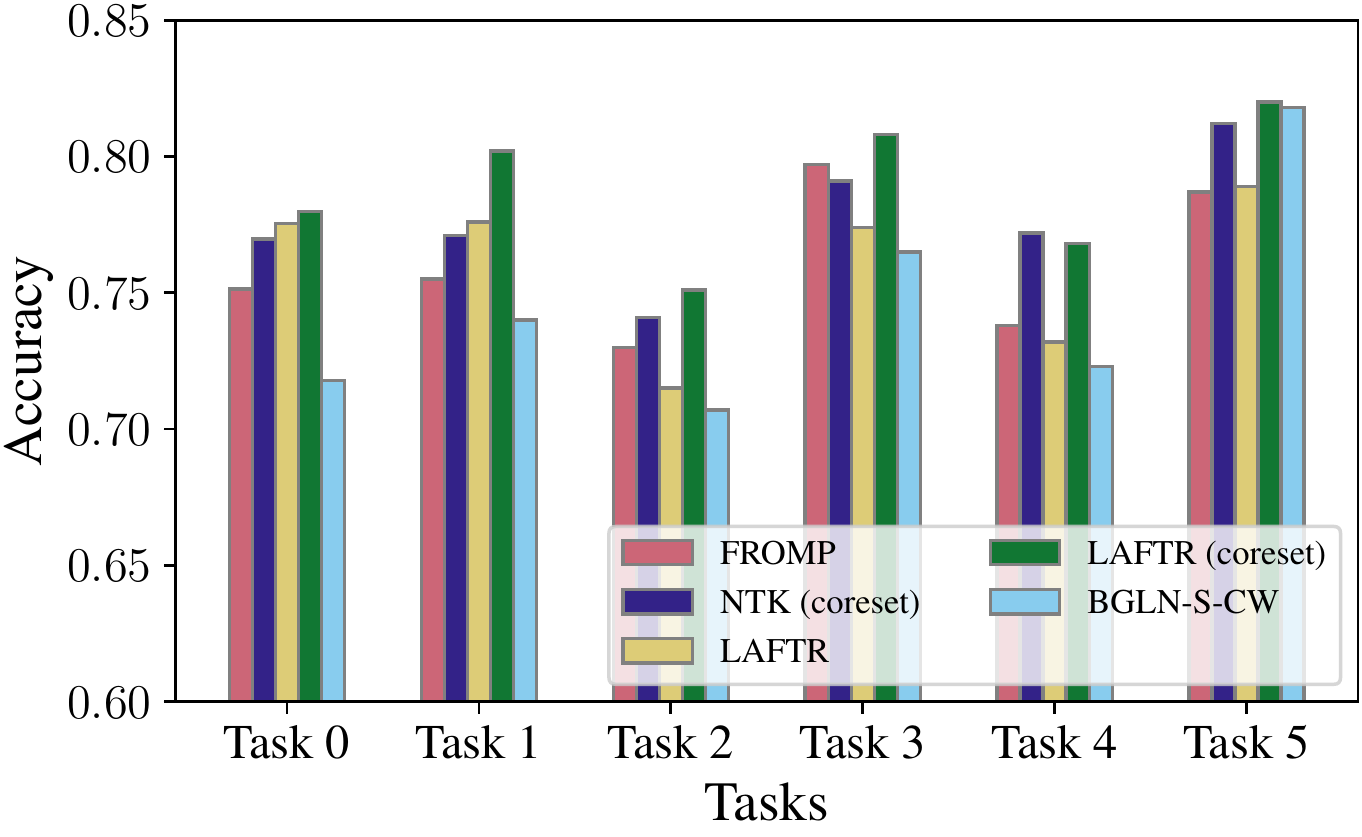}
    \caption{Comparison of task-wise accuracies on the Split CIFAR100 benchmark after training on all tasks is complete, for our methods, NTK and a nonparametric state-of-the-art method, FROMP.}
    \label{app:taskwise_fig}
\end{figure}

We show in Figure \ref{app:taskwise_fig} the task-wise accuracies on the Split CIFAR100 benchmark after training on all tasks is complete, for our methods (LAFTR, LAFTR (coreset) and BGLN-S-CW), NTK (coreset) and a nonparametric state-of-the-art method, FROMP. This draws a more fine-grained comparison and depicts the the benefits of LAFTR for each task.

To evaluate learning long sequences of tasks, we test our method’s performance on the extended, 11-task version of Split CIFAR100 (longer than the typical 6-task benchmark). The results for both versions of the task are shown in Table \ref{table:task_seq} for comparison.

\begin{table}[!t]
\caption{LAFTR (coreset) performance for different task sequence lengths.}
\begin{center}
\begin{tabular}{lcc}
\toprule
\multicolumn{1}{l}{\bf Number of tasks}  &\multicolumn{1}{c}{\bf Average Accuracy} &\multicolumn{1}{c}{\bf Backward Transfer} \\ 
\hline \\
\textbf{6} & $78.33 \pm 0.01$ & $- 0.73 \pm 0.10$ \\
\textbf{11} & $78.75 \pm 0.01$ & $- 1.59 \pm 0.47$ \\
\hline
\end{tabular}
\end{center}
\label{table:task_seq}
\end{table}

We find that performance of our method is maintained even for longer sequences of tasks. In this case, training on more tasks continually with LAFTR actually yields higher average accuracy. As we may expect, we observe a small decrease in the backward transfer performance, which still outperforms existing nonparametric SOTA methods.

\subsection{Ablation of information stored}
\label{mem_abl}
We present further results on the Split CIFAR100 task, with varying amounts of information stored and used in our method. Specifcally, Table \ref{table:coreset} summarizes the effect of coreset size on average accuracy and backward transfer for Split CIFAR100. It is also possible to generalize our class-conditional variant, which operates on $k=10$ clusters, to different number of clusters. We can create clusters by grouping classes and storing the required statistics for each cluster separately. As shown in Table \ref{table:clusters}, $k=1$ is sufficient to compete with existing nonparametric methods in this setting. As expected, as $k$ is increased and more dataset-level statistics are used, continual learning performance improves. 

\begin{table}[tbh]
\vspace{-0.2cm}
\caption{Effect of coreset size on Split CIFAR100 performance.}
\label{table:coreset}
\begin{center}
\begin{tabular}{lcc}
\toprule
\multicolumn{1}{l}{\bf Coreset Size}  &\multicolumn{1}{c}{\bf Average Accuracy} &\multicolumn{1}{c}{\bf Backward Transfer} \\ 
\hline \\
\textbf{20} & $73.26 \pm 0.01$ & $- 2.51 \pm 0.43$ \\
\textbf{50} & $75.63 \pm 0.01$ & $- 2.08 \pm 0.51$ \\
\textbf{100} & $76.70 \pm 0.01$ & $- 1.65 \pm 0.09$ \\
\textbf{200} & $78.33 \pm 0.01$ & $- 0.73 \pm 0.10$ \\
\hline
\end{tabular}
\end{center}
\end{table}

\begin{table}[tbh]
\caption{Effect of number of classwise clusters on Split CIFAR100 performance.}
\begin{center}
\begin{tabular}{lcc}
\toprule
\multicolumn{1}{l}{\bf k}  &\multicolumn{1}{c}{\bf Average Accuracy} &\multicolumn{1}{c}{\bf Backward Transfer} \\ 
\hline \\
\textbf{1} & $75.61 \pm 0.01$ & $- 1.93 \pm 0.59$ \\
\textbf{5} & $75.74 \pm 0.01$ & $- 1.58 \pm 0.50$ \\
\textbf{10} & $76.22 \pm 0.01$ & $- 1.45 \pm 0.63$ \\
\hline
\end{tabular}
\end{center}
\label{table:clusters}
\end{table}

\section{Influence Function Estimation}
\label{mis_ex}

\subsection{Experimental Details}

We used Concrete, Energy, Housing, Kinetics, and Wine datasets from the UCI collection~\citep{Dua:2019}. For all datasets, we normalized the training dataset to have a mean of 0 and a standard deviation of 1. We used a 2-hidden layer MLP with 128 hidden units and the base network was trained for 200 epochs with SGD and a batch size of 128. We performed hyperparameter searches over the learning rate in the range \{0.3, 0.1, 0.03, 0.01, 0.003, 0.001\} and selected the learning rate based on the validation loss. 

For each random data point selected, we optimized the PBRF objective for additional 20 epochs from the base network. The FSD term was computed stochastically with a batch size of 128. Similarly, both EWC and BGLN were trained with the same configuration but with the corresponding approximation to the FSD term.

\end{document}